\pdfoutput=1

\documentclass[11pt]{article}
\usepackage[most]{tcolorbox}

\usepackage[final]{acl} 
\usepackage{CJKutf8}
\usepackage[utf8]{inputenc}       
\usepackage[T1]{fontenc}          
\usepackage{inconsolata}
\usepackage{times}                
\usepackage{latexsym}             
\usepackage{booktabs}             
\usepackage{enumitem}             
\usepackage{algorithmic}          
\usepackage{microtype}            
\usepackage{tabularx}             
\usepackage{array}                
\usepackage{inconsolata}          
\usepackage{amsmath}              
\usepackage{xcolor}               
\usepackage{multirow}
\usepackage{amssymb}
\usepackage{CJKutf8}
\usepackage{xcolor}
\usepackage{tcolorbox}
\usepackage{multicol}
\usepackage{titlesec}
\usepackage{pifont} 
\newcommand{\cmark}{\ding{51}}  
\newcommand{\xmark}{\ding{55}}  
\usepackage{float}
\usepackage{makecell}
\usepackage{marvosym}   

\newcommand{\corremail}[1]{%
  \textsuperscript{\Letter}%
  \begingroup
    \renewcommand\thefootnote{}%
    \footnote{\Letter~Corresponding author (\href{mailto:#1}{#1}).}%
    \addtocounter{footnote}{-1}%
  \endgroup
}

\usepackage[ruled,vlined,linesnumbered]{algorithm2e}
\usepackage{etoolbox}
\robustify{\tcp}

\robustify{\ForEach}

\SetKwComment{Comment}{\texttt{>>}~}{}

\SetCommentSty{algcmt}   

\SetNlSty{algnl}{}{}     

\SetKwInOut{Input}{Input}
\SetKwInOut{Output}{Output}
\DontPrintSemicolon
\usepackage{amsfonts}

\usepackage{graphicx}             
\usepackage{tikz}                 
\usetikzlibrary{positioning, shapes.geometric, arrows.meta, calc}
\usepackage{pifont}
\usepackage{hyperref}
\usepackage[most]{tcolorbox}
\usepackage{titlesec}
\usepackage{CJKutf8}
\usepackage{natbib}
\makeatletter
\newif\ifavalon@was@twocol
\newenvironment{avalonscope}{%
  \par\begingroup
  \avalon@was@twocolfalse
  \if@twocolumn
    \avalon@was@twocoltrue
    \onecolumn
  \fi
  \raggedbottom
  \emergencystretch=2em
  \titlespacing*{\section}{0pt}{1.2ex plus .4ex minus .2ex}{0.8ex}
  \titlespacing*{\subsection}{0pt}{1.0ex plus .3ex minus .2ex}{0.6ex}
  \tcbset{
    enhanced, breakable, boxrule=0pt, sharp corners,
    before skip=6pt plus 2pt minus 1pt,
    after  skip=6pt plus 2pt minus 1pt
  }
}{%
  \par
  \ifavalon@was@twocol\twocolumn\fi
  \endgroup
}
\makeatother

\title{InMind: Evaluating LLMs in Capturing and Applying Individual Human Reasoning Styles}

\author{
  \textbf{Zizhen Li\textsuperscript{1,2}},
  \textbf{Chuanhao Li\textsuperscript{3}},
  \textbf{Yibin Wang\textsuperscript{4,2}},
  \textbf{Qi Chen\textsuperscript{5}},
  \textbf{Diping Song\textsuperscript{3}},
  \textbf{Yukang Feng\textsuperscript{1,2}}, \\
  \textbf{Jianwen Sun\textsuperscript{1,2}},
  \textbf{Jiaxin Ai\textsuperscript{6,2}},
  \textbf{Fanrui Zhang\textsuperscript{7,2}},
  \textbf{Mingzhu Sun\textsuperscript{1}},
  \textbf{Kaipeng Zhang\textsuperscript{3,2}\corremail{}} 
\\
\\
\textsuperscript{1}Nankai University,
\textsuperscript{2}Shanghai Innovation Institute
\textsuperscript{3}Shanghai AI Laboratory,
\textsuperscript{4}Fudan University, \\
\textsuperscript{5}Johns Hopkins University,
\textsuperscript{6}Wuhan University,
\textsuperscript{7}University of Science and Technology of China
\\[0.4em]
\texttt{lizizhen@mail.nankai.edu.cn}\\
\texttt{zhangkaipeng@pjlab.org.cn}
}

\begin{document}
\maketitle
\begingroup
  \renewcommand\thefootnote{\Letter}%
  \footnotetext{Corresponding author.}
\endgroup

\begin{abstract}
    LLMs have shown strong performance on human-centric reasoning tasks.
    While previous evaluations have explored whether LLMs can infer intentions or detect deception, they often overlook the individualized reasoning styles that influence how people interpret and act in social contexts.
    Social deduction games (SDGs) provide a natural testbed for evaluating individualized reasoning styles, where different players may adopt diverse but contextually valid reasoning strategies under identical conditions.
    To address this, we introduce \textbf{InMind}, a cognitively grounded evaluation framework designed to assess whether LLMs can capture and apply personalized reasoning styles in SDGs.
    InMind enhances structured gameplay data with round-level strategy traces and post-game reflections, collected under both Observer and Participant modes.
    It supports four cognitively motivated tasks that jointly evaluate both static alignment and dynamic adaptation.
    As a case study, we apply InMind to the game \textit{Avalon}, evaluating 11 state-of-the-art LLMs.\footnote{Dataset and code are released at \href{https://github.com/leroy9472/InMind}{\nolinkurl{https://github.com/leroy9472/InMind}}.}
    General-purpose LLMs, even GPT-4o frequently rely on lexical cues, struggling to anchor reflections in temporal gameplay or adapt to evolving strategies.
    In contrast, reasoning-enhanced LLMs like DeepSeek-R1 exhibit early signs of style-sensitive reasoning.
    These findings reveal key limitations in current LLMs’ capacity for individualized, adaptive reasoning, and position InMind as a step toward cognitively aligned human–AI interaction.
\end{abstract}

\section{Introduction}

\begin{figure}[h]
  \centering
  \includegraphics[width=\linewidth]{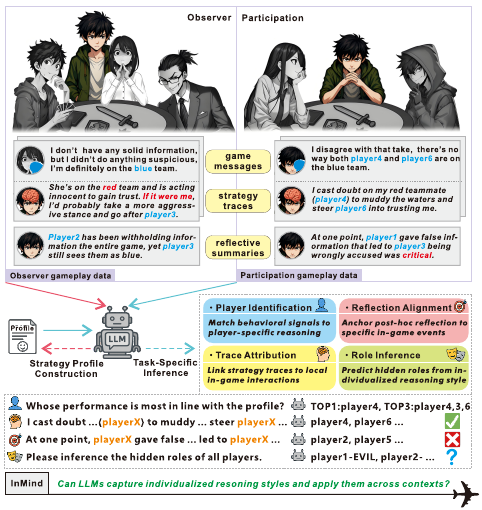}
\caption{\textbf{Overview of the InMind Framework.} The system constructs a subject-specific strategy profile from observer-mode data and applies it to participant-mode gameplay, supported by dual-layer annotations. Four cognitively motivated tasks assess the model’s ability to apply individualized reasoning styles.}
  \label{fig:framework-overview}
\end{figure}

Recent large language models (LLMs), such as DeepSeek-R1~\cite{deepseekai2025deepseekr1incentivizingreasoningcapability} and O1~\cite{openai2024openaio1card}, have demonstrated strong reasoning abilities across complex mathematical and scientific domains~\cite{chen2025reasoning}. Emerging research~\cite{Strachan2024ToM, Mittelstaedt2024SJT} further highlights their promising performance in human-centric tasks, including social commonsense inference, intention recognition, and belief attribution. Beyond these capabilities, recent studies suggest that LLMs may exhibit early signs of \textit{Theory of Mind} (ToM)—the ability to represent and reason about others' beliefs, desires, and intentions~\cite{saritas2025systematic, kim2025hypothesistom}. Understanding and evaluating such high-level cognitive traits is critical for advancing LLMs toward artificial general intelligence (AGI), and potentially, artificial superintelligence.

Existing benchmarks attempt to assess ToM-like reasoning through tasks such as intent classification~\cite{liu-etal-2024-interintent}, false-belief attribution~\cite{huang2024tomllm}, and multiple-choice social inference~\cite{seo-etal-2024-kocommongen}. However, these methods primarily target output plausibility or behavioral consistency, offering limited insight into underlying cognitive mechanisms, especially those that vary across individuals. In practice, different people often exhibit context-sensitive preferences in subjective scenarios and may arrive at similar conclusions via distinct reasoning trajectories~\cite{Otto2022Context,Charpentier2024Heterogeneity}. We refer to this as an \textit{individualized reasoning style}.

Social deduction games (SDGs) become an ideal evaluation scenario for internalizing and applying reasoning styles, where players must infer the hidden mental states of others and make strategic decisions accordingly~\cite{zhang2025multimindenhancingwerewolfagents,yoo2024finding}. Due to their dynamic, adversarial and individualized nature~\cite{feng2024surveylargelanguagemodelbased}, such settings require more than surface-level alignment: if an LLM cannot capture and adapt to a player’s individualized reasoning style, even plausible output may not support meaningful collaboration. Bridging this gap is essential for advancing ToM-inspired modeling of individual variation in reasoning, and for building LLMs capable of personalized, adaptive inference. We identify two key challenges: (1) how to capture and represent individualized reasoning processes, which may require structured interaction settings and cognitively meaningful annotations; (2) how to evaluate whether an LLM can apply a learned reasoning style in contextually adaptive ways, which calls for fine-grained, cognitively grounded tasks.

To meet these challenges, we propose \textbf{InMind}, a cognitively grounded evaluation framework designed to assess whether LLMs can internalize and apply individualized reasoning styles through SDGs. As illustrated in Figure~\ref{fig:framework-overview}, InMind introduces two complementary gameplay modes: \textit{Observer}, where a subject reasons passively from the perspective of another player without acting, and \textit{Participant}, where the subject actively engages in gameplay from their own perspective. This setup not only supports the natural capture of individualized reasoning, but also enables its application and evaluation in dynamic, interactive contexts.

Crucially, InMind integrates dual-layer cognitive annotations: (1) \textit{strategy traces}, which capture real-time reasoning signals such as belief updates, intention inference, and counterfactual thinking; and (2) \textit{reflective summaries}, offering post-hoc insights that contextualize key game events and assess the behaviors and intentions of other players. Leveraging these signals, InMind defines four cognitively motivated tasks to evaluate distinct aspects of individualized reasoning. 
(1) \textit{Player Identification} tests whether a model can recognize behavioral patterns that align with a specific reasoning style. 
(2) \textit{Reflection Alignment} assesses the model’s ability to ground abstract post-game reflections in concrete game behavior. 
(3) \textit{Trace Attribution} probes whether the model can simulate evolving, in-context reasoning across time. 
(4) \textit{Role Inference} evaluates whether the model can internalize reasoning styles to support belief modeling under uncertainty.

To concretely investigate these capabilities, we instantiate InMind within the popular social deduction game Avalon\footnote{\url{https://avalon-game.com/wiki/rules/}}, creating \textbf{InMind-Avalon}, a novel dataset comprising 30 full-session human gameplays annotated with detailed cognitive traces and reflective summaries. Our empirical analysis evaluates 11 state-of-the-art LLMs on Avalon-InMind and highlights several critical limitations: (1) Most models, including GPT-4o, heavily rely on superficial lexical patterns, failing to consistently infer deeper strategic intent; (2) Temporal alignment between reflective reasoning and specific in-game events remains challenging for nearly all evaluated models; (3) Dynamic adaptation of strategic reasoning based on evolving interactions is largely insufficient, indicating fundamental shortcomings in models’ capability for individualized reasoning. Nevertheless, we observe promising potential in certain models, such as DeepSeek-R1, suggesting possible avenues for improvement. Despite the inherent subjectivity in individualized annotations, these cognitively grounded traces and reflections effectively facilitate fine-grained tasks like hidden role identification, highlighting their practical utility for model training and evaluation.

In summary, our contributions are threefold:
(1) We introduce \textbf{InMind}, a cognitively grounded evaluation framework specifically designed to assess individualized reasoning and strategic adaptation of LLMs in dynamic social deduction scenarios;
(2) We release \textbf{InMind-Avalon}, a novel annotated dataset comprising 30 full-session human gameplay recordings enhanced with detailed cognitive annotations, including real-time strategy traces and reflective summaries;
(3)Through extensive evaluation of current state-of-the-art models, we identify critical limitations in temporally structured reasoning, adaptive strategy use, and individualized alignment.

We hope that InMind serves as a principled tool to guide future advances toward individualized, adaptive collaboration between humans and AI in socially rich, interactive environments.

\section{Related Work}

\subsection{Theory of Mind Reasoning in LLMs}

Recent research has shown that large language models (LLMs) increasingly demonstrate capabilities aligned with Theory of Mind (ToM), including false-belief attribution, intention recognition, and motivational reasoning~\cite{kim2025hypothesistom,saritas2025systematic}, which suggest that LLMs can approximate certain aspects of social cognition.
Several benchmarks have been introduced to evaluate these faculties, such as \textit{Social IQa}~\cite{sap-etal-2019-social}, \textit{KoCommonGEN}~\cite{seo-etal-2024-kocommongen}, and \textit{OpenToM}~\cite{xu2024opentomcomprehensivebenchmarkevaluating}, which typically use multiple-choice or context-driven tasks. More recent approaches incorporate dialog-based~\cite{yu2025persuasivetombenchmarkevaluatingmachine} and reinforcement learning settings~\cite{lu2025theorymindbenchmarksneed} to explore deeper social reasoning. 
However, most benchmarks focus on output plausibility and offer limited visibility into the reasoning process itself.
By contrast, the proposed InMind framework builds on these discussions and explicitly identifies two core limitations: the lack of temporal structure in evaluating reasoning over time, and the failure to distinguish between surface behavior and underlying cognition.

\subsection{Cognitive and Strategic Modeling in SDGs}

Social deduction games (SDGs), such as \textit{Avalon}, \textit{Werewolf}, and \textit{Among Us}, provide a dynamic and adversarial context for evaluating the strategic reasoning capabilities of LLMs~\cite{feng2024surveylargelanguagemodelbased,yoo2024finding,wu2024enhancereasoninglargelanguage}.
While several studies have utilized such environments to assess LLMs' performance in role identification, belief tracking, and deception detection~\cite{light2023avalonbenchevaluatingllmsplaying,zhang2025multimindenhancingwerewolfagents,chi2024amongagentsevaluatinglargelanguage}, most of them fail to provide structured representations of the cognitive processes involved in gameplay.
Although some efforts~\cite{stepputtis-etal-2023-long,liu-etal-2024-interintent} introduce temporal and intention-aware evaluations, they still fall short in providing comprehensive or individualized annotations that capture the evolving nature of reasoning within social deduction contexts.
In contrast, the InMind framework introduces cognitively annotated interactions across distinct gameplay modes, enabling the capture of individualized reasoning styles and supporting the evaluation of LLMs' adaptive capabilities in dynamic social settings.

\section{The InMind Framework}


We introduce InMind, a cognitively grounded framework for evaluating whether LLMs can internalize and apply individualized human reasoning styles. The framework is built on three key components:
(1) Structured Game Representation. InMind encodes gameplay using a structured representation that supports dual-perspective modeling and captures fine-grained cognitive annotations of each decision point, allowing for nuanced interpretation of reasoning behavior.
(2) Evaluation Protocol. InMind defines a protocol of four fine-grained evaluation tasks, designed to test both static alignment with human reasoning profiles and dynamic adaptability across varied gameplay contexts.
(3) InMind-Avalon: A Case Study. To demonstrate the framework in practice, we instantiate InMind in a case study based on the Avalon social deduction game, showcasing how the framework reveals personalized cognitive patterns and model behavior in complex, multi-agent settings.

\subsection{Structured Game Representation}
\label{sec:game-representation}

We represent each annotated game session as a structured tuple:
\begin{align}
\mathcal{G} = \langle \texttt{mode},\, \mathcal{A},\, \{E_z\}_{z \in \mathcal{Z}},\, \mathcal{F} \rangle.
\label{eq:game-structure}
\end{align}
Here, $\texttt{mode} \in \{\text{Observer}, \text{Participant}\}$ denotes the cognitive perspective under which the session was recorded.
The role assignment is defined as $\mathcal{A} = {(p_1, r_1), ..., (p_n, r_n)}$, where each player $p_i$ is assigned a hidden role $r_i$.
The game unfolds over rounds $\mathcal{Z} = {z_1, ..., z_m}$, with each round $z$ represented by an event tuple $E_z = \langle U_z, G_z, S_z \rangle$.
Moreover, $U_z$ contains all player utterances and system messages; $G_z$ records the observable game state (\textit{e.g.}, team proposals, votes, mission outcomes); and $S_z$ captures the subject’s real-time strategy trace, including their beliefs, intentions, and inferences.
Upon game completion, the subject provides a reflective summary $\mathcal{F}$ that articulates post-hoc reasoning, identifies pivotal moments, and evaluates other players’ behavior.
Together, the strategy trace $S_z$ and the reflective summary $\mathcal{F}$ constitute the dual-layer cognitive supervision central to InMind.

To simplify notation, we use superscripts $^o$ and $^p$ to indicate whether a variable comes from an \textit{Observer-mode} or \textit{Participant-mode} session. Both modes share the same data structure, the key distinction lies in the subject’s perspective.
In \textit{Observer-mode}, the subject does not take any actions but instead reasons from the perspective of a designated player.
This setup helps the model disentangle strategic reasoning from observed surface behavior.
In \textit{Participant-mode}, the subject actively engages in the game and provides annotations from their own point of view.

\subsection{Evaluation Protocol}
\label{sec:evaluation-protocol}

The evaluation protocol of InMind consists of two stages: Capturing Individual Reasoning Styles and Applying Profile in Adaptive Tasks, as depicted in Algorithm~\ref{alg:inmind}.
In Stage 1, it constructs a subject-specific reasoning profile to capture individual cognitive tendencies from \textit{Observer-mode} gameplay, independent of overt behavior.
In Stage 2, this profile is applied to a set of downstream reasoning tasks to evaluate the model’s ability to simulate and adapt to the subject’s decision-making style.

\subsubsection{Capturing Individual Reasoning Styles}
\label{sec:stage1-capture}



The goal of this stage is to derive a concise yet expressive profile $\mathcal{S}$ that captures the subject’s unique reasoning tendencies.
Rather than relying on explicit in-game actions, this process draws from \textit{Observer-mode} gameplay, in which the subject reasons aloud from the perspective of a designated player without taking any actions themselves.  
This design helps isolate cognitive patterns from behavioral noise, allowing for a more faithful reconstruction of individual reasoning styles.

To construct the profile, we apply a structured prompt (\texttt{ProfilePrompt}) over the subject’s full Observer-mode session $\mathcal{G}^o$.  
The prompt instructs the model to identify recurring interpretive strategies, decision heuristics, and evaluative criteria based on the subject’s commentary.  
The output is a free-form textual summary $\mathcal{S}$ that encapsulates how the subject tends to perceive, process, and respond to gameplay dynamics.  
This profile serves as a static cognitive signature, reused across all downstream evaluations to assess how well models internalize and adapt to human reasoning.

\subsubsection{Applying Profile in Adaptive Tasks}
\label{sec:stage2-apply}


In this stage, we assess whether a language model can leverage the reasoning profile $\mathcal{S}$, derived from Observer-mode data, to simulate the subject’s reasoning in novel gameplay contexts.
Given the profile $\mathcal{S}$ and a new Participant-mode session $\mathcal{G}^p$, the model is prompted to generate inferences that align with the subject’s cognitive style.

\begin{algorithm}[t]
\caption{InMind Evaluation Protocol}
\label{alg:inmind}
\KwIn{$\mathcal{G}^o$, $\mathcal{G}^p$, Tasks $\mathcal{T}$}
\KwOut{Results $\mathcal{R}$}

\tcp{Stage 1: Build subject-specific strategy profile }

$\mathcal{S} \gets \texttt{LLM}(\texttt{ProfilePrompt}(\mathcal{G}^o))$

\tcp{Stage 2: Apply profile to each task}
\ForEach{$T \in \mathcal{T}$}{
    \tcp{Construct task-specific context and targets}
    $(C, Y) \gets \texttt{ConstructTask}(\mathcal{G}^p, \mathcal{S}, T)$

    \tcp{Generate prediction via formatted prompt}
    $\hat{Y} \gets \texttt{LLM}(\texttt{FormatPrompt}(C, T))$

    \tcp{Compute evaluation result}
    $\mathcal{R} \gets \texttt{Eval}(\hat{Y}, Y)$
}
\end{algorithm}

\begin{table*}[t]
\centering
\small
\begin{tabular}{lcccccc}
\toprule
\textbf{Task} 
& \makecell[c]{\textbf{Profile} \\ $\mathcal{S}$} 
& \makecell[c]{\textbf{Game} \\ \textbf{Msgs} }
& \makecell[c]{\textbf{Strategy} \\ \textbf{Trace} $S_z$} 
& \makecell[c]{\textbf{Reflective} \\ \textbf{Summary}$\mathcal{F}$} 
& \makecell[c]{\textbf{Target}} 
& \makecell[c]{\textbf{Temporal} \\ \textbf{Mode}} \\
\midrule
Player Identification & $\checkmark$ & $\checkmark$ & $\checkmark$ & -- & Player ranking & Static \\
Reflection Alignment  & $\checkmark$ & $\checkmark$ & $\triangle$ & $\checkmark^\dagger$ & Player IDs & Static \\
Trace Attribution     & $\checkmark$ & $\checkmark$ & $\checkmark^\dagger$ & -- & Player IDs & Dynamic \\
Role Inference        & $\checkmark$ & $\checkmark$ & $\triangle$ & -- & Role assignment & Dynamic \\
\bottomrule
\end{tabular}
\caption{
Summary of task configurations in \textbf{InMind}. All tasks use the same subject-specific profile $\mathcal{S}$ and participant-mode gameplay as input. $\checkmark$ indicates components visible to the model; $\checkmark^\dagger$ marks fields where player identities are masked as prediction targets; $\triangle$ denotes optional inputs. Static tasks are presented as full-game contexts. Dynamic tasks reveal gameplay incrementally and require the model to summarize prior context, simulating human reasoning under limited memory and partial recall.
}
\label{tab:task-construction}
\end{table*}

\vspace{0.5em}

InMind transforms naturalistic annotations into structured prediction tasks by capitalizing on the inherently player-centric structure of social deduction games (SDGs).
Since reasoning in SDGs is organized around individual players and their interactions, most annotations are naturally linked to specific player IDs (\textit{e.g.}, P1–P6).
This structure allows cognitive traces and post-hoc reflections to be reformulated as clearly defined prediction problems.
To this end, we introduce four such tasks, including player identification, reflection alignment, trace attribution, and role inference.
All tasks are formulated as predictions over player identities or roles.
Their configurations are summarized in Table~\ref{tab:task-construction}.

\paragraph{(1) Player Identification.}
Given a subject-specific strategy profile $\mathcal{S}$ and a participant-mode session $\mathcal{G}^p$ (with all player identities anonymized), the model is tasked with identifying which player’s in-game behavior best aligns with the subject’s reasoning style. It produces a ranked list of candidates, and evaluation is based on Top-$k$ accuracy (typically $k=1,3$). This task measures static cognitive alignment between the profile and observed gameplay. Inputs include all player utterances and strategy traces, but exclude role assignments and game outcomes.

\paragraph{(2) Reflection Alignment.}

This task evaluates whether LLMs can ground high-level post-game reflections in concrete gameplay behavior. Each reflection $\mathcal{F}$, written by the subject, primarily comprises two forms of reasoning:   
(1) recalling a specific moment that affected the game’s trajectory (e.g., \textit{“Player 1 deceived Player 5 about mission sabotage”}), and  
(2) offering abstract evaluations of others based on global impressions (e.g., \textit{“Player 3 never voiced doubt, probably hiding something”}, which compresses multi-round silence into a single attribution).  
These reflections are typically temporally unanchored and do not reference specific rounds, posing challenges for grounding.
We mask player IDs mentioned in $\mathcal{F}$ and prompt the model to fill them in based on the full session $\mathcal{G}^p$ and the subject profile $\mathcal{S}$.
Performance is evaluated using exact-match accuracy, where a prediction is considered correct only if all masked player IDs in $\mathcal{F}$ are accurately recovered.

\paragraph{(3) Trace Attribution.}

This task assesses whether LLMs can simulate a subject’s evolving reasoning trajectory across rounds of gameplay. Each round-level trace $S_z$ reflects how the subject interprets recent events, including factual observations (e.g., \textit{“P3's last statement came across as overly eager”}), identity attributions (e.g., \textit{“I suspect P2 is Evil”}), and intended next moves (e.g., \textit{“I’ll accuse P4 next to test their reaction”}). To evaluate this capability, we mask all player IDs in $S_z$ and prompt the model to recover them incrementally(one round at a time), using only prior game context and the subject profile $\mathcal{S}$. Accuracy is measured by exact match. Unlike Task 2, this task requires real-time attribution and adjustment, testing whether models can follow the subject’s strategy as it unfolds.

\paragraph{(4) Role Inference.}
This task assesses whether LLMs can extend the subject’s reasoning style to perform belief modeling under uncertainty. As the game unfolds, the model must incrementally infer each player’s hidden role based on partial observations and evolving interactions.
At each round, it receives the current game history and outputs a full player-to-role mapping. Scoring is weighted toward later rounds to reflect the increasing availability of evidence. The task evaluates whether models can maintain consistency, adapt to new information, and infer the hidden roles

\subsection{InMind-Avalon: A Case Study}
\label{sec:avalon-dataset}

The InMind framework is instantiated in the six-player version of the social deduction game \textit{Avalon}, characterized by asymmetric hidden roles and the need for collaborative reasoning under uncertainty. The Good team comprises Merlin, Percival, and two Loyal Servants; the Evil team consists of Morgana and the Assassin. For detailed rules, please refer to Appendix~\ref{appendix:avalon-rules}.

We recruited 73 experienced players, one of whom was randomly selected to serve as the subject.
This player completed both observer-mode and participant-mode sessions, while other players were resampled per game to ensure strategic diversity. We then conducted all sessions via online voice chat in Mandarin Chinese to preserve authentic communication dynamics. Players frequently used game-specific expressions such as \textit{tiao pai}, \textit{dui tiao}, and \textit{chong piao}, which introduce additional reasoning challenges due to their implicit and context-dependent meanings. We transcribed all speech and comments verbatim, including disfluencies (\textit{e.g.}, pauses, hesitations), in order to preserve the real-time dynamics and interaction patterns of gameplay. To ensure annotation quality, three expert annotators accompanied the subject throughout and provided real-time guidance on producing round-level \textit{strategy traces} and post-game \textit{reflective summaries}. All annotations were reviewed for consistency. See Appendix~\ref{appendix:annotation-guidelines} and \ref{appendix:annotation-example} for details.

The resulting dataset is referred to as InMind-Avalon, which comprises 30 full game sessions (25 participant-mode, 5 observer-mode), totaling 884 player turns, 160 strategy traces, and 30 reflective summaries.
Each session lasts 20–25 minutes, with total gameplay exceeding 10 hours. Players are limited to 30 seconds per turn, yielding concise but tactically dense utterances. All canonical roles are well represented across varied team compositions. Notably, 22 games reached the final assassination phase, with Merlin correctly identified in only 41\% of cases, highlighting the difficulty of role inference. The subject’s team achieved a 56\% win rate in participant-mode games. Additional statistics are provided in Appendix~\ref{appendix:dataset-stats}.

Based on InMind-Avalon, we construct four structured evaluation tasks following the two-stage InMind protocol (Section~\ref{sec:evaluation-protocol}):  
(1) 25 player identification cases,
(2) 194 reflection alignment instances,  
(3) 791 trace attribution queries, and 
(4) 25 role inference sessions across 93 incremental rounds.

\section{Experiments}
\label{sec:experiments}

All experiments are conducted on the InMind-Avalon dataset (Section~\ref{sec:avalon-dataset}), following the two-stage protocol defined in Section~\ref{sec:evaluation-protocol}.  
In Section~\ref{sec:strategy-profile-case}, we analyze the generated strategy profiles from Stage 1.
Sections~\ref{sec:task1-player-identification} to \ref{sec:task4-role-inference} assess how effectively these profiles support downstream tasks, encompassing both static and dynamic reasoning. We evaluate 11 large language models (LLMs) under zero-shot settings. Detailed model specifications are provided in Appendix~\ref{appendix:models}, and the unified prompt format is described in Appendix~\ref{appendix:prompt}.

\subsection{Strategy Profile Analysis}
\label{sec:strategy-profile-case}

Each model begins by constructing a subject-specific strategy profile $\mathcal{S}$. We observe clear variation in profile quality and structure across models. GLM-4-9B ~\cite{glm2024chatglmfamilylargelanguage} typically produces vague personality summaries (\textit{e.g.}, describing the player as \textit{“logical and objective”} or \textit{“attentive to interpersonal interactions”}), while coherent, are generic and weakly grounded in gameplay. In contrast, DeepSeek-R1 ~\cite{deepseekai2025deepseekr1incentivizingreasoningcapability} generates multi-dimensional profiles that capture reasoning style, discourse tendencies, and adaptive strategies.
One such profile characterizes the subject as an \textit{“analytical assassin”}, who intentionally conceals analytical acuity, strategically employs probing questions to extract information, and even adopts Morgana’s perspective in Task 4 to infer how Percival was ultimately exposed.
This suggests that DeepSeek-R1 is able to extract abstract reasoning traits from observed strategic traces, going beyond surface-level linguistic cues.
A complete example is provided in Appendix~\ref{appendix:profile-example}.

\subsection{Player Identification}
\label{sec:task1-player-identification}

\paragraph{Evaluated Models}
Specifically, we select five general-purpose models varies in parameters (7B to 72B): Qwen2.5 ~\cite{qwen2025qwen25technicalreport}, Yi1.5 ~\cite{ai2025yiopenfoundationmodels}, GLM4 ~\cite{glm2024chatglmfamilylargelanguage}, InternLM2.5 ~\cite{wu2024internlm25stepproveradvancingautomatedtheorem} and GPT-4o ~\cite{openai2024gpt4ocard},
and three reasoning-enhanced models: DeepSeek-R1 ~\cite{deepseekai2025deepseekr1incentivizingreasoningcapability}, QwQ ~\cite{qwen2024qwq32b} and O3-mini ~\cite{openai2024o3o4} for evaluation.
Furthermore, we introduce a baseline model (``BERT Baseline'' in Table~\ref{tab:player-id-results}) for comparison, which does not use $\mathcal{S}$, but instead ranks candidates based on the cosine similarity between average-pooled StructBERT embeddings~\cite{wang2019structbertincorporatinglanguagestructures} of \( U_z^p \) and \( S_z^p \).

\paragraph{Setup}  

Each evaluated model receives $\mathcal{S}$, along with all player utterances \( U_z^p \) and strategy traces \( S_z^p \) , with roles and player identities withheld,
to output a ranked list of candidates whose behavior best matches the subject’s reasoning profile.
We report both Top-1 and Top-3 accuracy, along with BERT Match—the proportion of model predictions that align with the top-ranked candidate from the baseline.

\begin{table}[t]
\centering
\small
\begin{tabularx}{\linewidth}{@{}l *{3}{X}@{}}
\toprule
\textbf{Model} & \textbf{Top-1 \ \ \ Acc. $\uparrow$} & \textbf{Top-3 \ \ \ Acc. $\uparrow$} & \textbf{BERT Match $\downarrow$} \\
\midrule
BERT Baseline    & 0.160 & 0.480 & -- \\
\midrule
Qwen2.5-7B       & 0.168 & 0.416 & 0.200 \\
Qwen2.5-14B      & 0.168 & 0.496 & 0.208 \\
Qwen2.5-72B      & 0.208 & 0.544 & 0.272 \\
Yi1.5-9B         & 0.184 & 0.432 & 0.200 \\
Yi1.5-34B        & 0.104 & 0.456 & 0.160 \\
GLM4-9B          & 0.136 & 0.416 & 0.280 \\
InternLM2.5-20B  & 0.160 & 0.504 & 0.240 \\
GPT-4o           & 0.160 & \textbf{0.672} & 0.272 \\
\midrule
DeepSeek-R1      & \textbf{0.240} & 0.616 & \textbf{0.144} \\
QwQ              & 0.176 & 0.544 & 0.144 \\
O3-mini          & 0.200 & 0.576 & 0.288 \\
\bottomrule
\end{tabularx}
\caption{
Player identification accuracy.}
\label{tab:player-id-results}
\end{table}

\begin{figure}[h]
    \centering
    \includegraphics[width=0.95\linewidth]{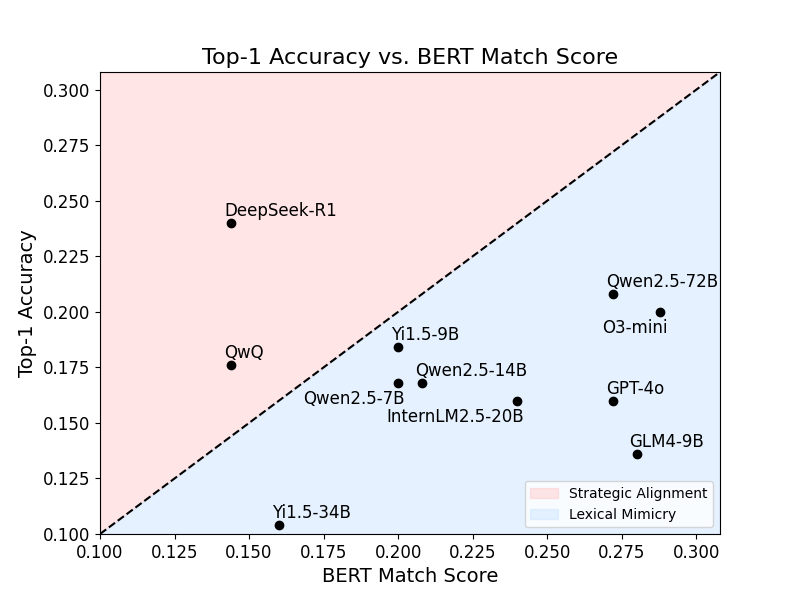}
\caption{Points above indicate stronger alignment with reasoning profiles beyond lexical similarity, while points below reflect greater similarity to surface-level patterns.
}
    \label{fig:reasoning-vs-mimicry}
\end{figure}

\paragraph{Results and Analysis}  
Table~\ref{tab:player-id-results} shows that overall Top-1 accuracy remains low across models, with most scores well below 0.20. Even Top-3 accuracy hovers around 0.5, close to chance level in a six-player setting. This highlights the inherent difficulty of identifying individualized reasoning styles from observable behavior, despite access to full gameplay data. Figure~\ref{fig:reasoning-vs-mimicry} further reveals the distinction between surface mimicry and deeper strategic alignment. Most models fall near or below the diagonal, indicating reliance on lexical similarity. DeepSeek-R1 stands out by achieving the highest Top-1 score (0.240) while maintaining the lowest BERT Match (0.144), suggesting more abstract reasoning-based alignment.

\subsection{Reflection Alignment}
\label{sec:task2-reflection-alignment}

\paragraph{Setup}  
Each evaluated model is given the subject’s profile $\mathcal{S}$, a participant-mode session $\mathcal{G}^p$ with masked player IDs in the reflection $\mathcal{F}$, and is asked to recover them.
We consider two evaluation settings:\textit{ Full Game Data}, which incorporates the strategy traces \( S_z^p \), and \textit{No Strategy Traces}, where these traces are omitted.
Accuracy is measured by exact match. Human expert performance is reported under the same conditions for reference.

\begin{figure}[h]
    \centering
    \includegraphics[width=0.95\linewidth]{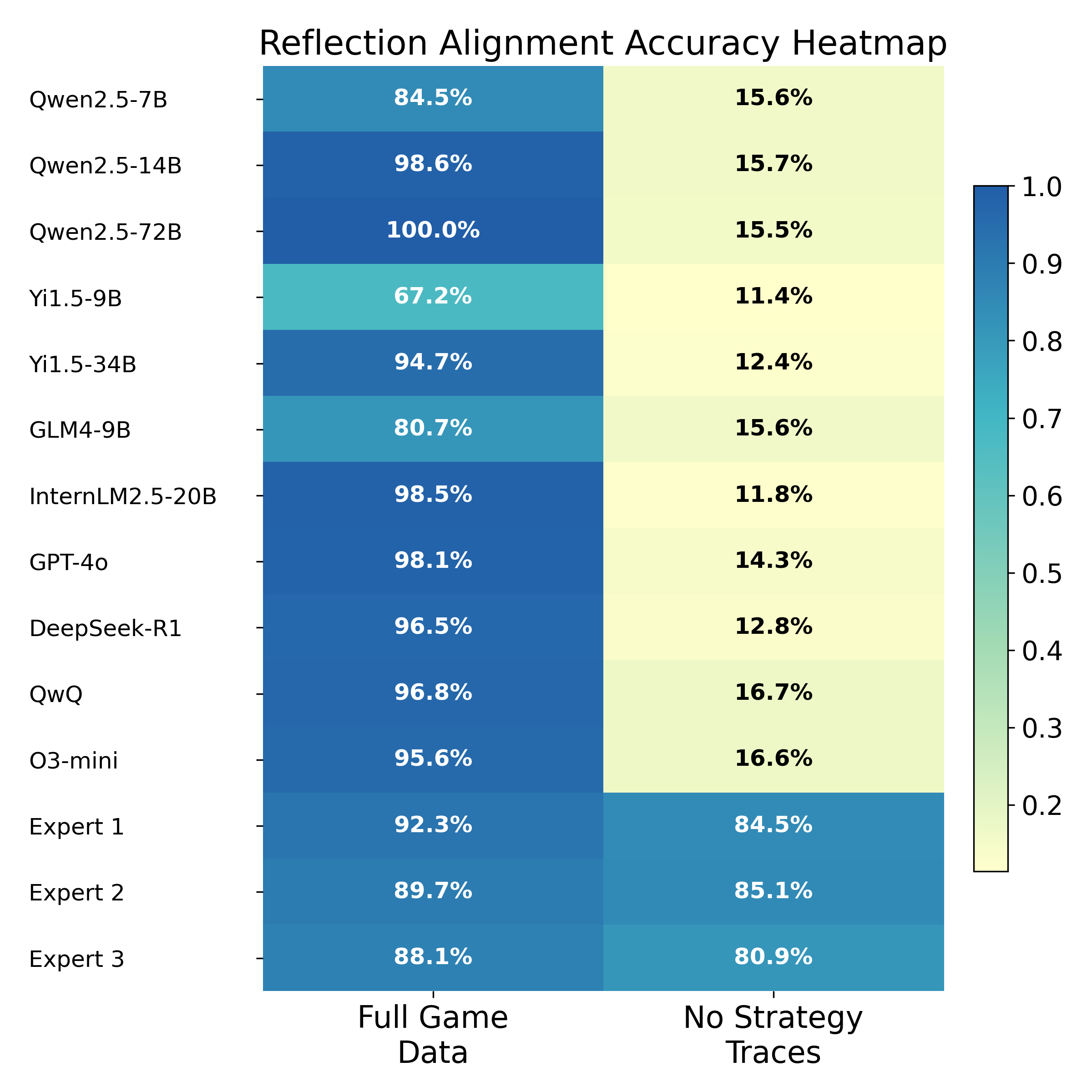}
    \caption{
Accuracy of player IDs prediction under two conditions: \textit{Full Game Data} (with access to strategy traces) and \textit{No Strategy Traces} (without traces). Results are shown for all models and three human experts.
}
    \label{fig:postgame-heatmap}
\end{figure}

\paragraph{Results and Analysis}
As shown in Figure~\ref{fig:postgame-heatmap}, models perform well when strategy traces are provided. Each trace is inherently linked to a specific round of gameplay, serving as a temporal anchor that ties reasoning to particular events. In this setting, the task effectively becomes a summarization of already structured signals. In contrast, when traces are withheld, accuracy drops sharply. Without clear temporal references, models struggle to associate abstract reflections with the appropriate moments and players in the game. This suggests a limited capacity for retrospective and context-aware reasoning, which the task is designed to evaluate. Human experts maintain high accuracy in both conditions, likely drawing on experience to reconstruct context even without explicit anchoring.

\subsection{Trace Attribution}
\label{sec:task3-trace-attribution}

\paragraph{Setup}  
At each round $z_i$, the model is given the subject profile $\mathcal{S}$ and all prior gameplay data $\mathcal{G}^p$ up to that point. It receives the current round’s utterances, game state, and a strategy trace $S_{z_i}$ with masked player IDs, and must predict the correct identifiers. The task proceeds incrementally, but each prediction is made independently, requiring the model to integrate and summarize all preceding context at each step. In the \textit{+Prior Trace} setting, the trace from the previous round $S_{z_{i-1}}$ is also provided as input. Evaluation is based on exact-match accuracy mentiond in Section \ref{sec:stage2-apply}.

\begin{table}[h]
\centering
\small
\begin{tabular}{lccc}
\toprule
\textbf{Model} & 
\makecell{\textbf{Base}\\\textbf{Accuracy $\uparrow$}} & 
\makecell{\textbf{+ Prior}\\\textbf{Trace $\uparrow$}} & 
\makecell{\textbf{Impact} \\ $\Delta$} \\
\midrule
Qwen2.5-7B         & 0.254 & 0.245 & -0.009 \\
Qwen2.5-14B        & 0.397 & 0.365 & -0.032 \\
Qwen2.5-72B        & 0.444 & 0.440 & -0.004 \\
Yi1.5-9B           & 0.206 & 0.197 & -0.009 \\
Yi1.5-34B          & 0.204 & 0.169 & -0.035 \\
GLM4-9B            & 0.241 & 0.224 & -0.017 \\
InternLM2.5-20B    & 0.226 & 0.215 & -0.011 \\
GPT-4o             & 0.440 & 0.448 & +0.008 \\
\midrule
DeepSeek-R1        & \textbf{0.503} & \textbf{0.517} & \textbf{+0.014} \\
QwQ                & 0.437 & 0.454 & +0.017 \\
O3-mini            & 0.268 & 0.281 & +0.013 \\
\bottomrule
\end{tabular}
\caption{
Trace attribution accuracy with and without access to the prior round’s strategy trace. The final column reports the performance impact ($\Delta$), where positive values indicate successful adaptation to evolving context.
}
\label{tab:trace-attribution-results}
\end{table}

\paragraph{Results and Analysis}  
As shown in Table~\ref{tab:trace-attribution-results} and Figure~\ref{fig:strategy-reflection-delta}(see Appendix~\ref{appendix:trace-impact}), most models show little to no benefit from accessing the previous trace $S_{z_{i-1}}$, and some even decline in accuracy. This indicates difficulty in leveraging prior reasoning to inform current predictions. Rather than building on evolving beliefs, models tend to treat each round as an isolated instance, reflecting limited integration of temporal strategy context. These results highlight a core limitation in dynamic attribution: current LLMs struggle to track and reproduce individualized reasoning styles over time, making it difficult to maintain coherent, round-by-round inference.

\subsection{Role Inference}
\label{sec:task4-role-inference}

\paragraph{Setup}  
We task the model with inferring the hidden roles of all players from the perspective of a designated subject, based on observed gameplay. To examine how contextual cues influence role inference, we vary three factors: (1) whether the prompt adopts a first-person or third-person perspective, (2) whether the subject’s round-level strategy traces ($S_z$) are provided, and (3) whether the subject’s own role is revealed. These conditions define four prompting modes (A–D), summarized in Table~\ref{tab:role-inference-matrix}.


Each setting is evaluated under two criteria: strict scoring and relaxed scoring.
The former requires the model to accurately identify all five canonical roles, while the latter simplifies the task by grouping roles into three broader categories: \textit{Informed Good} (Merlin, Percival), \textit{Uninformed Good} (Loyalists), and \textit{Evil} (Morgana, Assassin).

\begin{table}[h]
\centering
\small
\begin{tabular}{lccc}
\toprule
\textbf{Mode} & \textbf{Perspective} & \textbf{Trace Access} & \textbf{Role Known} \\
\midrule
A & First-person  & \cmark & \cmark \\
B & First-person  & \xmark & \cmark \\
C & Third-person  & \xmark & \cmark \\
D & Third-person  & \xmark & \xmark \\
\bottomrule
\end{tabular}
\caption{
Prompting modes for role inference. All configurations are tested under both strict (exact role) and relaxed (role group) scoring.
}
\label{tab:role-inference-matrix}
\end{table}

\begin{figure}[t]
    \centering
    \includegraphics[width=\linewidth]{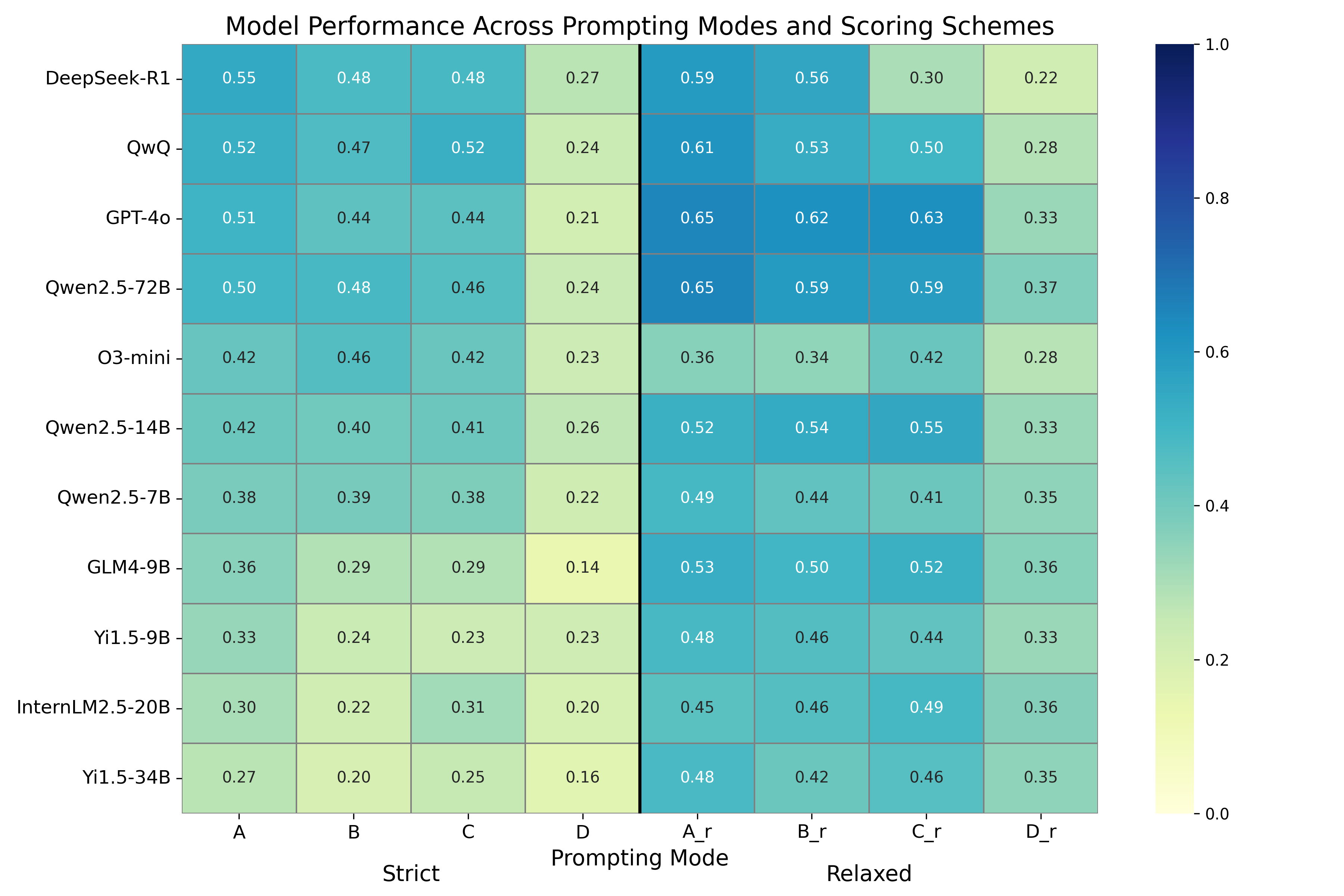}
\caption{
Each cell shows average accuracy under one of four prompting modes (A–D), evaluated using both strict (exact role) and relaxed (role group) criteria. Stronger models appear toward the top. 
}
    \label{fig:model-mode-heatmap}
\end{figure}

\paragraph{Results and Analysis}
Figure~\ref{fig:model-mode-heatmap} and Figure~\ref{fig:rating-mode-barplot}(See ~\ref{appendix:prompt-mode-role-analysis}) show that providing access to strategy traces $S_z$ improves model performance across all prompting configurations. Mode A achieves the highest scores under both strict and relaxed conditions, indicating that subjective annotations, even when potentially biased or incomplete, can support more effective role inference. In contrast, shifting the prompt perspective from first-person (Mode B) to third-person (Mode C) yields similar results. This suggests that LLMs do not show a clear benefit from observer-style prompting, and we do not observe the “outsider sees more of the game” phenomenon commonly associated with human reasoning. While strict role identification remains challenging for most models, their performance under relaxed scoring demonstrates emerging potential for collaborative inference in social reasoning settings.

We also investigate how the subject’s role (e.g., Merlin or Percival) shapes the demands placed on reasoning models, as different roles require distinct inference strategies. This supplementary analysis, presented in Appendix~\ref{appendix:role-by-initial-role}, helps assess whether models can adapt to varying cognitive perspectives.

\section{Conclusion}
InMind offers a novel, cognitively grounded framework for evaluating whether large language models can internalize and apply individualized reasoning styles in complex, interactive settings. By leveraging the structured dynamics of social deduction games and enriching gameplay data with dual-layer annotations from multiple perspectives, InMind enables fine-grained assessment of strategic and adaptive reasoning. Our accompanying dataset, InMind-Avalon, and comprehensive model evaluation reveal key limitations in current LLMs’ adapting to individualized human reasoning styles. We believe this framework paves the way for more personalized, socially aware AI systems and invites further exploration into cognitive modeling and adaptive collaboration.

\section*{Limitations}
While InMind is designed as a general framework for evaluating individualized reasoning in social deduction games, the current implementation focuses exclusively on the Avalon setting. Although Avalon captures many of the strategic and cognitive elements characteristic of SDGs, our experiments do not yet explore other game environments with different social structures or interaction patterns. In future work, we plan to expand both the range of tasks and the scale of the dataset to include additional games such as \textit{Blood on the Clocktower}, \textit{The Resistance} and \textit{Werewolf}, thereby enriching the diversity of reasoning styles and dynamics.

Given the inherently subjective nature of individualized reasoning, the annotation process unavoidably reflects annotator preferences and interpretation. Although expert annotators were involved throughout to guide and standardize the process, variation and bias are difficult to eliminate entirely. Expanding the dataset in both size and diversity will help mitigate such subjectivity and improve the robustness and generalizability of InMind.

Moreover, the core ideas behind InMind extend beyond games and hold potential for broader domains such as multi-agent collaboration, negotiation, and human-AI teaming, where personalized, context-sensitive reasoning is essential. Supporting real-time, multi-agent dynamics will be an important step toward evaluating and enhancing LLMs in more complex, socially situated environments.

\section*{Ethical considerations}
The gameplay data used in this study was collected and provided by a  author, who conducted the sessions in accordance with institutional ethical guidelines. All participants were informed about the nature and purpose of the data collection and gave their consent prior to participation. They were also provided with opportunities to ask questions and seek clarification, ensuring informed and voluntary participation. No personally identifiable information was included in the dataset, and all data is used strictly for academic research purposes in compliance with relevant data protection regulations.

\section*{Acknowledgement}
This paper is supported by the National Key R\&D Program of China No.2022ZD0160102. The computations in this research were performed using the computing platform of the Shanghai Innovation Institute. We thank the anonymous reviewers for their valuable and constructive feedback, which greatly improved this work.

\bibliography{custom}

\appendix
\clearpage
\section{Game Setting and Terminology}
\label{appendix:avalon-rules}
\subsection{ Avalon Setup and Roles}

In the InMind-Avalon dataset, all sessions follow the 6-player variant of \textit{Avalon}, featuring asymmetric roles and hidden identities. The game consists of two opposing teams: \textbf{Good team (4 players)}: Merlin, Percival, and two Loyal Servants; \textbf{Evil team (2 players)}: Morgana and the Assassin. Table~\ref{tab:avalon-vision} summarizes the visibility structure among roles.

\begin{table*}[h]
\centering
\begin{tabular}{@{}lccc@{}}
\toprule
\textbf{Role} & \textbf{Team} & \textbf{Knows Others} & \textbf{Seen By} \\
\midrule
Merlin         & Good & Morgana, Assassin        & – \\
Percival       & Good & Merlin and Morgana (ambiguous) & – \\
Loyal Servant  & Good & –                       & – \\
Morgana        & Evil & Assassin                & Merlin, Percival \\
Assassin       & Evil & Morgana                 & Merlin \\
\bottomrule
\end{tabular}
\caption{Vision and asymmetric knowledge structure in Avalon (6-player setting).}
\label{tab:avalon-vision}
\end{table*}

\vspace{0.5em}
\noindent

\vspace{0.5em}
\noindent \textbf{Merlin.} Knows the identities of all evil players, but must avoid revealing this knowledge. Typically hides among loyal players and subtly steers team decisions.

\noindent \textbf{Percival.} Is informed that two players are Merlin and Morgana but does not know who is who. Aims to distinguish the real Merlin and support them covertly.

\noindent \textbf{Loyal Servant.} Has no knowledge of roles. Relies purely on reasoning and observation. Must avoid overexposing suspicion to protect Merlin’s cover.

\noindent \textbf{Morgana.} Pretends to be Merlin to confuse Percival and manipulate good players. May mislead, mimic, or publicly claim roles to create chaos.

\noindent \textbf{Assassin.} Knows Morgana’s identity. Remains hidden, tracks likely Merlin candidates, and ultimately attempts to assassinate Merlin if the good side wins.

\subsection{Game Flow}

Avalon proceeds through a series of up to 5 missions, each comprising three stages:

\vspace{0.4em}
\noindent \textbf{Team Formation.}  
Each round begins with a designated leader proposing a team for the mission. Players then speak in turn to express their stance, analyze the game state, and (optionally) make role claims. Finally, all players vote on the proposed team. If the vote passes by majority, the team proceeds. Otherwise, leadership passes to the next player. If five consecutive votes fail, the evil team wins by default (forced vote rule).

\vspace{0.4em}
\noindent \textbf{Mission Execution.}  
Selected team members privately vote “success” or “fail.” Good players must vote “success,” while evil players may vote either way. A mission fails if at least one player chooses “fail.” Mission team sizes by round are fixed: 2-3-4-3-4 (total of 5 missions). Team composition influences deduction dynamics.

\vspace{0.4em}
\noindent \textbf{Victory Conditions.}  
The first team to win three missions wins the game. If the good team wins three missions, the Assassin is given one opportunity to identify Merlin. If correct, the evil team wins; otherwise, the good team wins. In special cases, the evil team may “knife” Merlin early if both evil players agree, resulting in an immediate win upon success.

\begin{CJK*}{UTF8}{gbsn}
\subsection{On Preserving Chinese Terminology}

All data in InMind-Avalon is collected from live Chinese-language gameplay and annotation. We intentionally preserve core in-game terms in Chinese for two reasons: Many phrases (\textit{e.g.}, “跳派”, “踩”, “拱”) carry cultural or strategic nuance not captured by direct translation. Retaining these terms enhances reproducibility and supports linguistic fidelity during model training and evaluation. table~\ref{tab:avalon-terms} provides key terminology used throughout the dataset, with English explanations.

\end{CJK*}

\vspace{0.6em}
\begin{CJK*}{UTF8}{gbsn}
\begin{table*}[htbp]
\centering
\begin{tabular}{@{}lp{0.18\linewidth}p{0.68\linewidth}@{}}
\toprule
\textbf{中文术语} & \textbf{English Gloss} & \textbf{Explanation} \\
\midrule
跳派       & claim Percival     & Declare oneself as Percival to influence team dynamics. \\
对跳       & counterclaim       & Multiple players claim the same role to cause confusion. \\
拇指       & thumbs             & Percival sees two “thumbed” players: Merlin and Morgana. \\
红 / 蓝     & red / blue          & Evil / good team members. \\
踩 / 拱     & accuse / endorse   & Lower or raise suspicion on other players. \\
上车 / 车下 & on/off team        & Selected or not selected for the mission team. \\
冲票       & force vote         & Vote yes to prevent mission stalemate. \\
刀梅林     & knife Merlin       & Evil team's final attempt to guess Merlin's identity. \\
挡刀 / 躲刀 & take / avoid knife & Sacrifice or mislead to protect Merlin. \\
自爆       & self-destruct      & Deliberately reveal one’s evil role to disrupt alignment. \\
\bottomrule
\end{tabular}
\caption{Glossary of common Chinese gameplay terms used in Avalon.}
\label{tab:avalon-terms}
\end{table*}
\end{CJK*}

\section{Annotation Guidelines and Examples}
\label{appendix:annotation-guidelines}

\begin{CJK*}{UTF8}{gbsn}
\subsection{Subject Annotation Manual}

This manual was provided to the designated subject to guide consistent, cognitively grounded annotations across all sessions in the InMind-Avalon dataset. Annotations are divided into two types: \textbf{strategy traces} written at the end of each round, and a \textbf{reflective summary} written after each game concludes. Subjects were instructed to write freely in Chinese, using Avalon-specific expressions, and to focus on their internal reasoning process from the perspective of their assigned role.

\vspace{0.5em}
\textbf{该指南提供给实验对象，用于指导其在 InMind-Avalon 数据集中完成一致且具备认知深度的标注。标注包含两类：每轮结束后的 \textbf{策略轨迹}，以及整局游戏结束后的 \textbf{反思总结}。被试使用中文自然表达，鼓励使用阿瓦隆常用术语，从所扮演角色的视角出发记录真实思考过程。}

\subsubsection*{Strategy Trace ($S_z$)}

At the end of each game round, the subject should briefly summarize their thought process based on the observed state and prior discourse. Each trace should reflect the evolving beliefs, suspicions, intentions, and situational inferences from the subject’s perspective.

\vspace{0.3em}
\textbf{每轮游戏结束后，记录你在该轮结束时的思考与判断。内容应涵盖你对当前局势的分析、身份的猜测、未来打算等。}

Suggested components include:

\begin{itemize}
    \item \textbf{回顾局势变化或关键行为}  
    “上一轮 2 明明信 4，现在突然踩，感觉在试图洗身份。”  
    \textit{(“In the last round, 2 trusted 4, but suddenly accused him now — feels like they’re trying to reset their image.”)}

    \item \textbf{当前发言和投票的推理判断}  
    “5 说不想上 3，但又跟票同意进队，很不一致。”  
    \textit{(“Player 5 said they didn’t want 3 in the team but still voted yes — very inconsistent.”)}

    \item \textbf{身份推测与信任关系}  
    “目前我觉得 1 是派，2 像民，3 有点像莫甘娜。”  
    \textit{(“I think 1 is Percival, 2 feels like a loyal, and 3 might be Morgana.”)}

    \item \textbf{下一轮打算或备选策略}  
    “下一轮我想跟 2 或 4 组，试探一下他们的反应。”  
    \textit{(“Next round I’ll try teaming with 2 or 4 to see how they respond.”)}

    \item \textbf{可能存在的反常信号或混淆因素}  
    “3 一直跟风踩 5，但 5 投得很正常，怀疑 3 在带节奏。”  
    \textit{(“3 keeps piling on 5, but 5’s behavior seems fine — I suspect 3 is trying to stir the pot.”)}

    \item \textbf{情绪波动与犹豫}（可选）  
    “现在我也有点乱了，感觉大家都在演。”  
    \textit{(“Honestly I’m getting confused — everyone’s putting on an act.”)}
\end{itemize}

\subsubsection*{Reflective Summary ($\mathcal{F}$)}

After each game, the subject writes a high-level reflection summarizing how their beliefs and reasoning evolved, what moments were pivotal, and how they evaluate others' actions in hindsight.

\vspace{0.3em}
\textbf{游戏结束后，从整体上回顾自己的策略、关键时刻、误判与判断依据，以及对其他玩家的评价。}

Suggested components include:

\begin{itemize}
    \item \textbf{游戏中的关键节点回顾}  
    “第 2 轮投票 1 给了反对票，我没在意，现在看来很反常。”  
    \textit{(“In Round 2, 1 voted no on a team — I ignored it at the time, but now it feels very suspicious.”)}

    \item \textbf{某些行为的事后推断}  
    “3 原来是梅林，怪不得一直默默带我走正解。”  
    \textit{(“Turns out 3 was Merlin — now I understand why they subtly guided me to the right team.”)}

    \item \textbf{哪些人演得像民/狼/梅林，为何？}  
    “4 很积极踩人，看起来像民，其实是莫甘娜，演技不错。”  
    \textit{(“4 was aggressively calling people out, looked like a loyal, but turned out to be Morgana — good acting.”)}

    \item \textbf{本局中的误判与经验教训}  
    “我太相信 2 了，他说话一直很像忠，但其实是刀。”  
    \textit{(“I trusted 2 too much — he always sounded loyal but was actually the assassin.”)}

    \item \textbf{若重来一次，会如何调整策略？}  
    “下次我不会再无脑信跟我一条线的人，要再看一轮再判断。”  
    \textit{(“Next time I won’t blindly trust people who agree with me — I’ll wait one more round to assess.”)}
\end{itemize}

\end{CJK*}

\begin{CJK*}{UTF8}{gbsn}
\subsection{Expert Review Manual}

This guideline was provided to expert reviewers for validating and refining the strategy traces and reflective summaries generated by the subject. Reviewers were instructed to focus on coherence, consistency, and cognitive depth, while preserving the natural language style and spontaneity of the subject’s reasoning.

\vspace{0.5em}
\textbf{本指南用于指导专家评审者对被试生成的策略轨迹与反思总结进行校验与修正，确保标注具有一致性、合理性与认知有效性，同时保留其自然语言表达和个体思维风格。}

\subsubsection*{General Instructions}

\begin{itemize}
    \item Preserve the subject’s personal reasoning style, language choices, and terminology (\textit{e.g.}, “跳派”, “踩”, “狼”).
    \item Avoid correcting grammar or expression unless it affects clarity or logic.
    \item Focus on the \textbf{alignment between annotations and game events} (dialogue, votes, team changes).
    \item If key reasoning steps are missing, use comments to prompt clarification from the subject (\textit{e.g.}, “Why did you start trusting Player 3 here?”).
    \item Ensure all names, numbers, and references (\textit{e.g.}, player IDs) match the game context.
\end{itemize}

\subsubsection*{Reviewing Strategy Traces ($S_z$)}

Each trace should reflect turn-level reasoning grounded in the current round and previous context. Reviewers should:

\begin{itemize}
    \item Check that the trace reflects the correct round context (\textit{e.g.}, reflects current voting or player behavior).
    \item Ensure belief updates and intentions are logically coherent.
    \item Mark vague statements for elaboration if they lack justification.
    \item Maintain ambiguity where natural — overconfidence is not required.
\end{itemize}

\vspace{0.3em}
\textbf{示例（原始）:} “这轮我觉得 4 是梅林。”  
\textbf{建议修改:} “4 发言不像民，而且跟我上一轮分析方向一致，我觉得他可能是梅林。”  
\textit{(“4 didn’t sound like a loyal, and his statements matched my earlier deductions — I suspect he’s Merlin.”)}

\subsubsection*{Reviewing Reflective Summaries ($\mathcal{F}$)}

Reflective summaries should capture global reasoning across the session. Reviewers should:

\begin{itemize}
    \item Ensure the subject reflects on at least 2–3 major moments (\textit{e.g.}, turning points, team shifts, hidden role reveals).
    \item Encourage inclusion of both accurate and mistaken judgments.
    \item Highlight inconsistencies between reflection and trace progression (\textit{e.g.}, player previously trusted is now judged negatively without explanation).
    \item Ask for clarification when post-hoc evaluations are too vague or unsupported.
\end{itemize}

\vspace{0.3em}
\textbf{示例（原始）:} “1 表现很像狼。”  
\textbf{建议修改:} “1 第三轮突然强跳派，还踩了 4，但第四轮又改口说 4 很像好人，这种转变让我觉得很狼。”  
\textit{(“1 suddenly claimed to be Percival in Round 3 and attacked 4, but in Round 4 praised 4 — this shift made me suspicious.”)}

\subsubsection*{Final Review Actions}

\begin{itemize}
    \item Use inline comments for clarification requests or proposed edits.
    \item If a trace is severely off-context, suggest partial rewriting with explicit reasoning.
    \item After review, compile a list of trace/summary entries needing subject clarification.
    \item All final edits must be approved by at least two reviewers and the subject before inclusion.
\end{itemize}

\noindent
\textbf{Note:}  
Expert review is not intended to “correct” the player’s thinking, but to ensure that the annotations faithfully represent in-game cognition and can be reliably interpreted for model supervision and evaluation.

\vspace{0.5em}
\textbf{说明：}  
专家评审旨在保障标注内容的认知逻辑性和上下文一致性，而非强行规范语言或统一推理风格。评审者应协助被试清晰表达其真实推理，并对关键缺失信息提供引导性建议。

\end{CJK*}

\section{Dataset Statistics and Visualizations}
\label{appendix:dataset-stats}

To assess the behavioral diversity and annotation coverage of our InMind-Avalon dataset, we present several quantitative distributions.

\begin{figure}[h]
    \centering
    \includegraphics[width=0.8\linewidth]{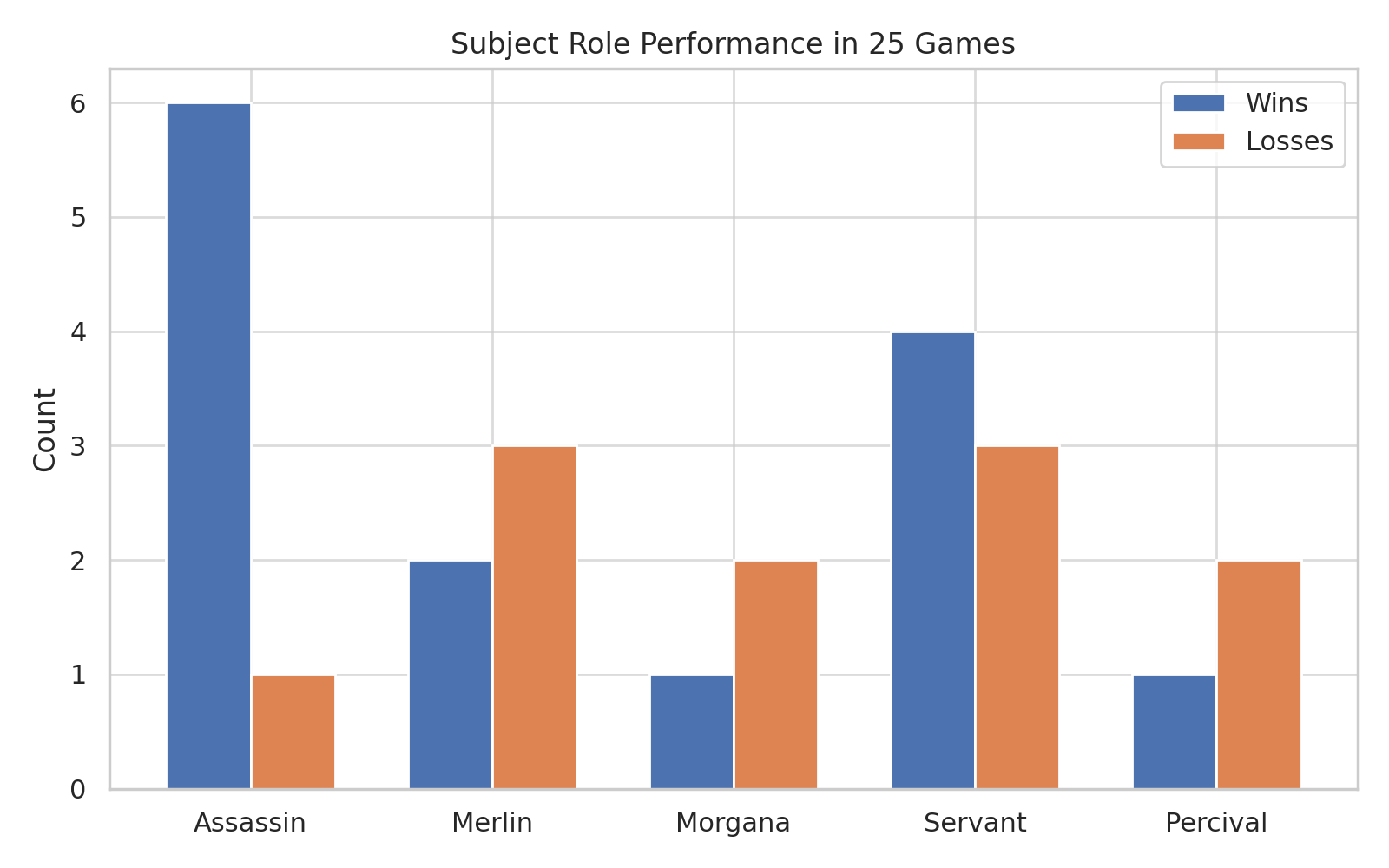}
    \caption{Win/loss counts per role in the 25 participant-mode games. The subject played all five canonical roles.}
    \label{fig:winloss}
\end{figure}

\begin{figure}[h]
    \centering
    \includegraphics[width=0.8\linewidth]{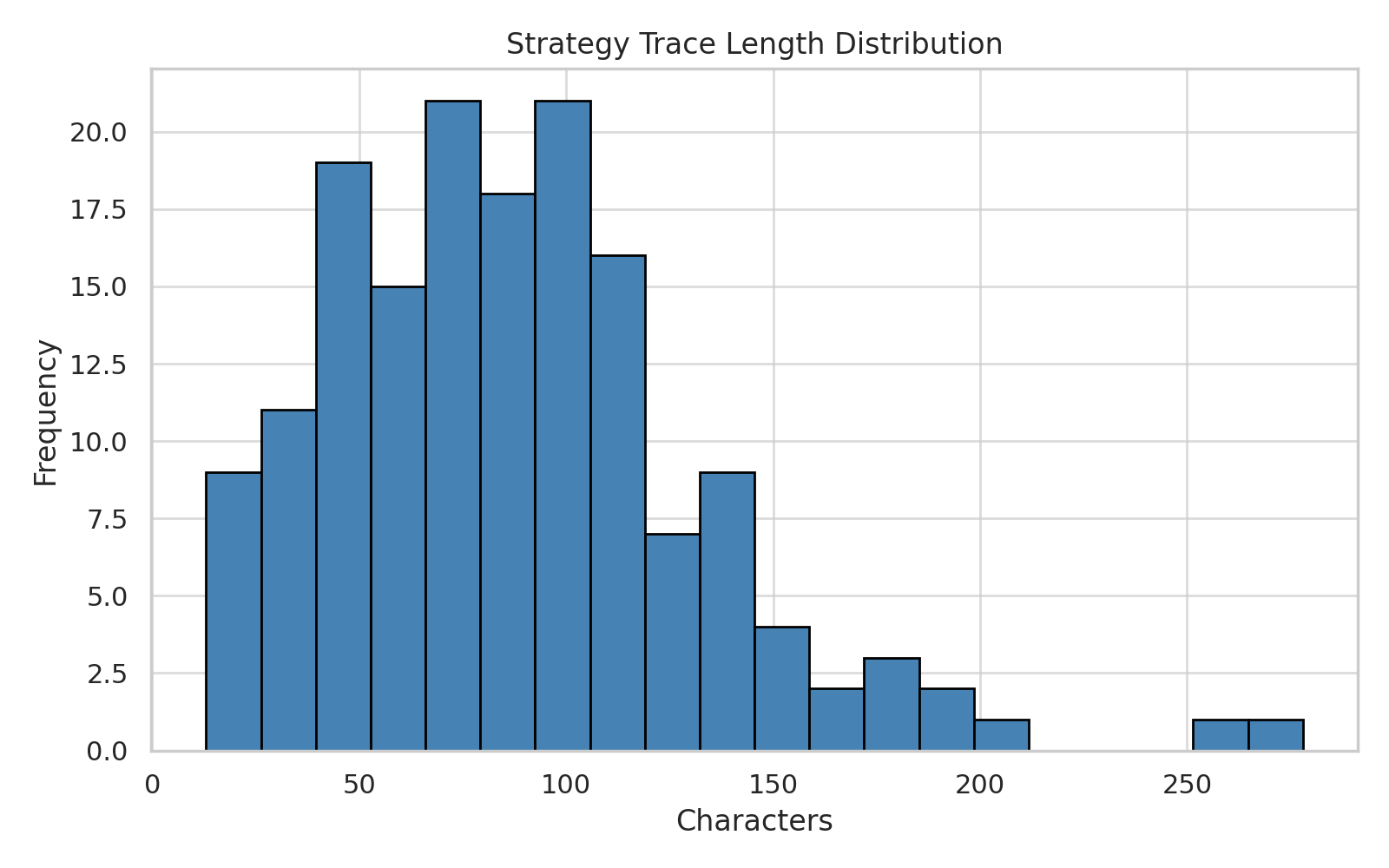}
    \caption{Distribution of strategy trace lengths (in characters). The average length is 87.4 characters, with standard deviation 45.1.}
    \label{fig:trace-len}
\end{figure}

\begin{figure}[h]
    \centering
    \includegraphics[width=0.8\linewidth]{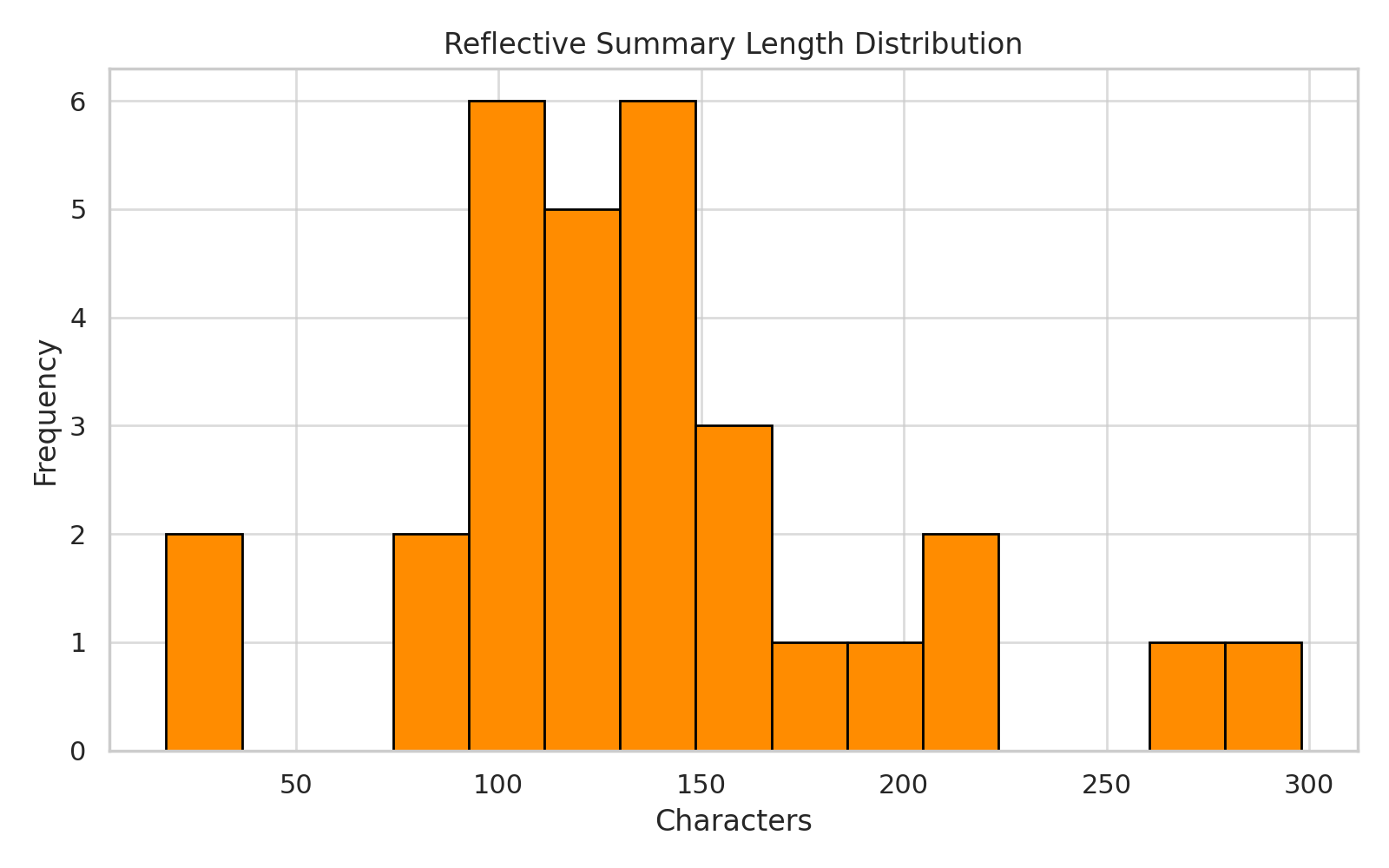}
    \caption{Distribution of reflective summary lengths. The average is 135.4 characters, standard deviation 56.3.}
    \label{fig:review-len}
\end{figure}

\begin{table}[ht]
\centering
\begin{tabular}{lcc}
\toprule
\textbf{Role} & \textbf{Wins} & \textbf{Losses} \\
\midrule
Assassin         & 6 & 1 \\
Merlin           & 2 & 3 \\
Morgana          & 1 & 2 \\
Loyal Servant    & 4 & 3 \\
Percival         & 1 & 2 \\
\bottomrule
\end{tabular}
\caption{Win/loss record across 25 Participant-mode sessions by role.}
\label{tab:role-results}
\end{table}

Of the 22 games that reached the assassination phase (including 7 early assassination attempts), the Evil side succeeded in identifying Merlin in 41\% of cases. Strategy traces span 160 entries across games; summaries vary widely in style and granularity.

\section{Model Details}
\label{appendix:models}

{
\setlength{\tabcolsep}{2pt}
\begin{table}[htbp]
\centering
\small
\begin{tabular}{lccc}
\toprule
\textbf{Model} & \textbf{Parameters} & \textbf{Access} & \textbf{Source} \\
\midrule
GPT-4o             & Unknown        & API       & OpenAI  \\
Qwen2.5-72B        & 72B            & Local     & Alibaba  \\
Qwen2.5-14B        & 14B            & Local     & Alibaba  \\
Qwen2.5-7B         & 7B             & Local     & Alibaba  \\
DeepSeek-R1        & 671B            &API     & DeepSeek \\
Yi-1.5-34B         & 34B            & Local     & 01.AI  \\
Yi-1.5-9B          & 9B             & Local     & 01.AI \\
GLM-4-9B           & 9B             & Local     & Zhipu.AI  \\
InternLM2.5-20B    & 20B            & Local     & Shanghai AI Lab  \\
QwQ                & 32B            & Local       & Alibaba  \\
O3-mini            & Unknown             & API     & OpenAI   \\
\bottomrule
\end{tabular}
\caption{
Summary of evaluated models, their parameter sizes, access methods, and sources.  
“API” models are queried via official endpoints; “Local” models are run with fixed weights in offline inference.
}
\label{tab:model-details}
\end{table}
}

\section{Prompt Templates}

\label{appendix:prompt}
To support reproducibility and interpretability of all evaluation tasks introduced in Section~\ref{sec:evaluation-protocol}, we provide detailed prompt templates in both Chinese and English. Each prompt follows the two-stage protocol described in Algorithm~\ref{alg:inmind}, with the first stage producing a strategy profile from observation-mode games and the second stage performing task-specific inference using this profile.

All prompts were originally used in Mandarin Chinese to preserve pragmatic fidelity. Below we present bilingual representations of representative prompts. Task-specific variants and full inputs are provided as supplementary material.

\addcontentsline{toc}{section}{Appendix E: Prompt Templates}

\begin{CJK*}{UTF8}{gbsn}  
\noindent

\begin{table*}[t]
\centering
\small
\begin{tcolorbox}[enhanced, sharp corners, boxrule=0.4pt, colback=gray!3, colframe=black, width=\textwidth, title=E.1~Strategy Profile Construction Prompt（策略画像构建）]
\textbf{System Prompt（中文）}:\par
你是一名精通阿瓦隆桌游的分析专家，擅长透过玩家的行为和发言推测玩家的推理逻辑、人物风格和行动策略。你的任务是基于用户提供的旁观游戏记录，精准地总结用户在旁观时展现的推理风格、发言倾向以及分析局势的方法，为之后识别用户实际参与游戏时的发言特点建立准确的用户画像。

\medskip
\textbf{System Prompt (English)}:\par
You are an expert analyst in the Avalon board game. Your task is to infer a reasoning style, player persona, and behavioral tendencies based on the user’s observation of others’ gameplay. Your summary should help construct an accurate profile that can be used to identify this user’s future behavior when actively participating in the game.

\medskip
\textbf{User Prompt（中文）}:\par
以下是我旁观某位玩家时记录的一局游戏数据，包含我的分析和评价：

\verb|{REFLECTION_SEGMENT}|

结合此前你已经帮我总结过的：
\verb|{ACCUMULATED_SUMMARY}|

请基于之前的总结以及上述新的游戏数据，重新生成一个更全面、更细致的『综合玩家画像』，该画像应包括但不限于：

- 我的推理风格（如侧重逻辑推理、发言细节推敲、人际互动观察等）
- 我可能的发言特征（如发言长度、用词风格、信息分享的方式、阵营表态的倾向）
- 我对游戏局势的常规分析策略（如喜欢从哪些信息判断敌我关系、如何处理模糊信息、如何引导队友）

要求精准、细致地进行归纳总结，以便下一步准确推测实际游戏中的我。

\medskip
\textbf{User Prompt (English)}:\par
Below is a reflection record I made while observing another player’s game, containing my analysis and commentary:

\verb|{REFLECTION_SEGMENT}|

You have already helped me summarize:
\verb|{ACCUMULATED_SUMMARY}|

Based on this new gameplay observation and previous summaries, please generate an updated and comprehensive strategy profile. Your summary should include (but is not limited to):

- My reasoning style (\textit{e.g.}, logical deduction, detail-oriented analysis, social interaction tracking)
- My possible language or expression features (\textit{e.g.}, verbosity, tone, information disclosure, alliance signaling)
- My habitual strategy for analyzing game state (\textit{e.g.}, what signals I prioritize, how I handle ambiguity, how I influence others)

Please provide a precise and detailed characterization to guide future inference about my behavior when I actively play.
\end{tcolorbox}
\caption{Bilingual system and user prompt for constructing a strategy profile from observer-mode reflections. Placeholders such as \texttt{\{REFLECTION\_SEGMENT\}} and \texttt{\{ACCUMULATED\_SUMMARY\}} are replaced with session-specific content.}
\label{tab:prompt-profile}
\end{table*}
\end{CJK*}

\begin{table*}[t]
\centering
\small
\begin{CJK*}{UTF8}{gbsn}
\begin{tcolorbox}[enhanced, sharp corners, boxrule=0.4pt, colback=gray!3, colframe=black, width=\textwidth, title=E.2~Player Identification Prompt（玩家识别任务）]
\textbf{System Prompt（中文）}:\par
你是一名精通阿瓦隆桌游的分析专家，擅长透过玩家的发言和行为推测玩家的真实身份。此前你已经详细总结了用户旁观时的推理风格、人物特点和典型发言倾向，建立了一个准确的人物画像。

你当前的\textbf{唯一任务}是：\\
根据用户的『综合玩家画像』，精准地在实际游戏数据中，通过分析玩家的发言特点、用词习惯和推理策略，识别哪位玩家是用户本人。

请严格按照以下输出要求：
\begin{itemize}
    \item 仅输出玩家编号（player1 至 player6）
    \item 不要输出任何额外解释或说明
    \item 输出最可能的玩家 (Top1) 和最可能的前三玩家 (Top3)，严格以JSON结构给出。
\end{itemize}

\medskip
\textbf{System Prompt (English)}:\par
You are an expert in Avalon social deduction analysis. Based on the user's prior reasoning profile and speaking habits, your sole task is to identify the player most likely to be the user from a new game session.

Strictly follow the format below:
\begin{itemize}
    \item Output only player indices (player1 to player6)
    \item Do not provide any explanation
    \item Return Top-1 and Top-3 predictions in \texttt{JSON} format
\end{itemize}

\medskip
\textbf{User Prompt（中文）}:\par
此前你已为我精准总结了我的『综合玩家画像』：
\verb|{final_summary}|

以下是一局我实际参与的游戏数据：
- \texttt{player\_message}：包含本局游戏所有玩家的发言记录（含我本人）  
- \texttt{strategy}：包含我对当前局势的内部分析和我的发言策略等细节  
\verb|{new_game_prompt}|

你的\textbf{唯一任务}是根据上述游戏数据，判断在编号为 player1 至 player6 的玩家中，哪位玩家最可能是我。

严格按照以下格式输出：
\begin{verbatim}
{
    "top1": "playerX",
    "top3": ["playerX", "playerY", "playerZ"]
}
\end{verbatim}

注意：
\begin{itemize}
    \item 请严格遵循 JSON 格式，仅输出玩家编号
    \item 排名应准确反映与『综合玩家画像』的匹配程度
\end{itemize}

\medskip
\textbf{User Prompt (English)}:\par
Previously you have summarized my personalized reasoning profile:  
\verb|{final_summary}|

Now you are given a full gameplay session that I participated in:  
- \texttt{player\_message}: utterances from all players, including me  
- \texttt{strategy}: my internal reasoning and actions (anonymized)  
\verb|{new_game_prompt}|

Your sole task is to determine which of player1 through player6 is most likely me.

Return:
\begin{verbatim}
{
    "top1": "playerX",
    "top3": ["playerX", "playerY", "playerZ"]
}
\end{verbatim}

Only output the JSON. No explanations.
\end{tcolorbox}
\end{CJK*}
\caption{Prompt template for Player Identification. The model is asked to match observed player behavior to a prior reasoning profile and return the most likely player index in structured format.}
\label{tab:prompt-id}
\end{table*}

\begin{table*}[t]
\centering
\small
\begin{CJK*}{UTF8}{gbsn}
\begin{tcolorbox}[enhanced, sharp corners, boxrule=0.4pt, colback=gray!3, colframe=black, width=\textwidth, title=D.3~Reflection Alignment Prompt（复盘对齐任务）]

\textbf{System Prompt（中文）}:\par
你是一个擅长分析阿瓦隆游戏的大模型。  
根据以下提示，你的任务是为我的复盘（my\_review）中的每个 \texttt{[MASK\_x(...)]} 填充数字或数字组合，并严格遵守以下格式要求：  
1. 对于 \texttt{[MASK\_x(1digit)]}，请用单个数字表示该位置的玩家编号。  
2. 对于 \texttt{[MASK\_x(Ndigits)]}，请用多个数字组成一个数字组合，数字按升序排列，表示多位玩家标号。  
3. 你需要完整地为每个 \texttt{[MASK\_x(...)]} 提供一个对应的数字或数字组合映射行。  
4. 不要输出多余的映射行，不要输出多余的其他解释或文字。

\medskip
格式要求：  
\begin{verbatim}
[MASK_x(1digit)] => y
[MASK_x(Ndigits)] => yz
\end{verbatim}

\medskip
\textbf{System Prompt (English)}:\par
You are a large language model skilled in analyzing Avalon gameplay.  
Your task is to fill in each \texttt{[MASK\_x(...)]} in my post-game reflection (my\_review) with player indices, following strict format rules:  
1. For \texttt{[MASK\_x(1digit)]}, fill a single player number.  
2. For \texttt{[MASK\_x(Ndigits)]}, fill an ascending sequence of player numbers.  
3. Provide exactly one mapping line per mask.  
4. Do not output any extra mappings or explanations.

\medskip
Required format:  
\begin{verbatim}
[MASK_x(1digit)] => y
[MASK_x(Ndigits)] => yz
\end{verbatim}

\medskip
\textbf{User Prompt（中文）}:\par
此前你已经帮我总结了：我旁观时的推理风格和人物特点以及参与游戏时可能的复盘内容，  
\verb|{final_summary}|

以下是一局我参与的游戏数据\verb|{new_game_prompt}|：
\vspace{0.5em}

\begin{tabularx}{\textwidth}{>{\bfseries}lX}
My Info & 玩家在游戏开始时获得的初始角色信息 \\[0.5em]
Tasks Overview (Round by Round) & \\
\quad player\_message & 玩家及系统的发言记录（包括我的发言） \\
\quad strategy & 我对每一轮游戏局势的分析和策略说明 \\[0.5em]
Final User Info & 游戏结束后每位玩家的真实角色身份
\end{tabularx}

\vspace{0.5em}

你的\textbf{唯一任务}：根据总结和游戏数据，推测每个我对整局游戏的复盘内容中被MASK的玩家编号，并输出映射行。  
- [My Review]：\verb|{masked_msg}|

\textbf{要求}：  
1. 每个 \texttt{[MASK\_x(...)]} 都需要填写 1~6 之间的数字编号，按升序排列。  
2. 格式要求严格：\texttt{[MASK\_x(1digit)] => y} 或 \texttt{[MASK\_x(Ndigits)] => yz}  
3. 不要输出多余的映射行，仅填写实际需要的映射行，不要解释。  

\medskip
\textbf{User Prompt (English)}:\par
Previously, you have summarized my reasoning style and typical reflection content:  
\verb|{final_summary}|

Below is a gameplay session I participated in\verb|{new_game_prompt}|: 

\vspace{0.5em}

\begin{tabularx}{\textwidth}{>{\bfseries}lX}
My Info & Initial roles I received \\[0.5em]
Tasks Overview (Round by Round) & \\
\quad player\_message & Utterances from players and the system (including mine) \\
\quad strategy & My round-by-round analyses and speaking strategies \\[0.5em]
Final User Info & Ground-truth roles of all players
\end{tabularx}
\vspace{0.5em}

Your sole task: fill in each \texttt{[MASK\_x(...)]} in my post-game reflection with player numbers based on the summary and game data, outputting only the mappings.  
- [My Review]: \verb|{masked_msg}|

Requirements:  
1. Each \texttt{[MASK\_x(...)]} should be filled with numbers 1–6, in ascending order.  
2. Format strictly as: \texttt{[MASK\_x(1digit)] => y} or \texttt{[MASK\_x(Ndigits)] => yz}  
3. Output only necessary mappings, no explanations.
\end{tcolorbox}
\end{CJK*}
\caption{Prompt template for Reflection Alignment task. The model fills masked player references in the user's post-game reflection with correct player indices.}
\label{tab:prompt-reflection}
\end{table*}

\begin{table*}[t]
\centering
\small
\begin{CJK*}{UTF8}{gbsn}
\begin{tcolorbox}[enhanced, sharp corners, boxrule=0.4pt, colback=gray!3, colframe=black, width=\textwidth, title=D.4~Trace Attribution Prompt（策略轨迹归属任务）]

\textbf{System Prompt（中文）}:\par
你是一个擅长分析阿瓦隆游戏的大模型。  
你会根据我的推理风格，分析每一轮游戏数据并根据要求输出内容。请确保所有的输出严格遵守格式要求，尤其是替换映射的格式。

你将接收到我提供的游戏数据，要求你对其进行分析，按照以下方式输出：

1. 游戏数据的总结（Content部分）以 \texttt{=== Content ===} 开头，按顺序总结任务的内容。  
2. 策略中被MASK的玩家编号以及对应的映射行，以 \texttt{=== Replacements ===} 开头，并按照 \texttt{MASK\_x(Ndigits) => abc...} 的格式输出。请确保仅输出实际需要的映射行，且格式无误。  
对于 \texttt{[MASK\_x(1digit)]}，请用单个数字表示该位置的玩家编号，对于 \texttt{[MASK\_x(Ndigits)]}，请用多个数字组成一个数字组合，数字按升序排列，表示多位玩家标号。

要求：  
- 格式要求严格：输出中的替换映射应遵循 \texttt{[MASK\_x(1digit)] => m} 或 \texttt{[MASK\_x(Ndigits)] => abc...}。  
- 不输出多余的映射行，仅填写实际需要的映射行，注意映射行一定要完整。  
- 不输出任何其他内容。

\medskip
\textbf{System Prompt (English)}:\par
You are a large language model skilled in analyzing Avalon gameplay.  
You will analyze incremental gameplay data according to my reasoning style and produce output as required. Please strictly follow the formatting rules for replacement mappings.

You will receive gameplay data from me and are asked to produce:

1. A summary of the game data (Content section) beginning with \texttt{=== Content ===}, sequentially summarizing the task content.  
2. Replacement lines for masked player indices in my strategy (strategy) with \texttt{=== Replacements ===}, formatted as \texttt{MASK\_x(Ndigits) => abc...}. Output only required mapping lines with exact format.  
For \texttt{[MASK\_x(1digit)]}, fill a single player number; for \texttt{[MASK\_x(Ndigits)]}, fill an ascending sequence of player numbers.

Requirements:  
- Replacement mappings must strictly follow \texttt{[MASK\_x(1digit)] => m} or \texttt{[MASK\_x(Ndigits)] => abc...} format.  
- Do not output extra mappings, only necessary ones.  
- No additional text or explanation.

\medskip
\textbf{User Prompt（中文）}:\par
此前你已经帮我总结了：我旁观游戏时表现出的推理风格和复盘策略，  
\verb|{final_summary}|

下面我会逐步地向你提供一局我实际参与的新的游戏数据，以下是你对此前游戏数据的总结，  
\verb|{data_conc}|

下面是新的部分游戏数据，  
\verb|{new_game_prompt}|

你有两个任务：  
1. 请你总结全部的游戏数据，包括关键的玩家发言和游戏状态等，以 \texttt{=== Content ===} 开头。  
2. 请你根据我的推理风格，输出我的策略（strategy）中被MASK的玩家编号，并输出映射行，以 \texttt{=== Replacements ===} 开头。

注意事项：  
1. 格式要求严格：\texttt{[MASK\_x(1digit)] => m} 或 \texttt{[MASK\_x(Ndigits)] => abc...}。  
2. 不要输出多余的映射行，仅填写实际需要的映射行。  
3. 即使不确定玩家编号，也要输出完整的映射行。  
4. 不要输出其他任何多余的内容。

****注意：两个任务一定要分别用\texttt{=== Content ===}和\texttt{=== Replacements ===}开头。****

\medskip
\textbf{User Prompt (English)}:\par
Previously, you have summarized my reasoning style and reflection strategies:  
\verb|{final_summary}|

I will now incrementally provide new gameplay data from a session I participated in, along with your previous summary,  
\verb|{data_conc}|

Below is the new partial gameplay data,  
\verb|{new_game_prompt}|

You have two tasks:  
1. Summarize all gameplay data, including key player utterances and game states, starting with \texttt{=== Content ===}.  
2. Based on my reasoning style, output the masked player numbers in my strategy with replacement lines starting with \texttt{=== Replacements ===}.

Notes:  
1. Replacement mappings must strictly follow \texttt{[MASK\_x(1digit)] => m} or \texttt{[MASK\_x(Ndigits)] => abc...} format.  
2. Output only required mappings, no extras.  
3. Output complete mappings even if uncertain.  
4. No extra text.

****Note: Two tasks must be clearly separated by \texttt{=== Content ===} and \texttt{=== Replacements ===}.****
\end{tcolorbox}
\end{CJK*}
\caption{Prompt template for Trace Attribution task. The model outputs a summary of gameplay and player ID replacements for masked indices in the user's strategy.}
\label{tab:prompt-trace}
\end{table*}

\begin{table*}[t]
\centering
\small
\begin{CJK*}{UTF8}{gbsn}
\begin{tcolorbox}[enhanced, sharp corners, boxrule=0.4pt, colback=gray!3, colframe=black, width=\textwidth, title=D.5~Role Inference Prompt（角色推断任务）]

\textbf{System Prompt（中文）}:\par
你是一个阿瓦隆游戏角色推断专家。你熟悉阿瓦隆游戏的所有角色和规则，并能够根据给定信息准确推断玩家角色。请根据后续提供的游戏数据，输出6位玩家的角色，要求如下：  
- 1名[梅林]  
- 1名[莫甘娜]  
- 1名[派西维尔]  
- 1名[刺客]  
- 2名[忠臣]

请严格按照以下格式输出，不要添加其他任何内容或说明：  
\texttt{=== Roles ===}  
Player 1: [角色]  
Player 2: [角色]  
Player 3: [角色]  
Player 4: [角色]  
Player 5: [角色]  
Player 6: [角色]

\medskip
\textbf{System Prompt (English)}:\par
You are an Avalon role inference expert. You are familiar with all roles and rules of Avalon and can accurately infer player roles based on given information. Please output the roles of the six players according to the forthcoming gameplay data, with the following requirements:  
- 1 Merlin  
- 1 Morgana  
- 1 Percival  
- 1 Assassin  
- 2 Loyal Servants

Strictly follow the output format below without any extra content or explanation:  
\texttt{=== Roles ===}  
Player 1: [Role]  
Player 2: [Role]  
Player 3: [Role]  
Player 4: [Role]  
Player 5: [Role]  
Player 6: [Role]

\medskip
\textbf{User Prompt（中文）}:\par
下面我会逐步地向你提供一局我实际参与的新的游戏数据，以下是你对此前游戏数据的总结，  
\verb|{data_conc}|

下面是新的部分游戏数据，  
\verb|{new_game_prompt}|

你有两个任务：  
1. 请你总结全部的游戏数据，包括关键的玩家发言和游戏状态等，以 \texttt{=== Content ===} 开头。  
2. 请你根据玩家和系统发言，结合我的角色信息（代入我的角色视角进行分析），输出每位玩家可能的角色，以 \texttt{=== Roles ===} 开头。

格式要求如下：  
Player 1: [角色]  
Player 2: [角色]  
Player 3: [角色]  
Player 4: [角色]  
Player 5: [角色]  
Player 6: [角色]

****注意两个任务一定要分别以\texttt{=== Content ===}和\texttt{=== Roles ===}开头。****

\medskip
\textbf{User Prompt (English)}:\par
I will incrementally provide a new game session data I participated in, along with your previous summary,  
\verb|{data_conc}|

Here is the new partial game data,  
\verb|{new_game_prompt}|

You have two tasks:  
1. Summarize all game data including key player utterances and game state with \texttt{=== Content ===} header.  
2. Based on player and system utterances, and my role information (analyzing from my perspective), output the likely role for each player starting with \texttt{=== Roles ===}.

Format requirements:  
Player 1: [Role]  
Player 2: [Role]  
Player 3: [Role]  
Player 4: [Role]  
Player 5: [Role]  
Player 6: [Role]

****Note that the two tasks must begin with \texttt{=== Content ===} and \texttt{=== Roles ===} respectively.****
\end{tcolorbox}
\end{CJK*}
\caption{Prompt template for Role Inference task. The model outputs a summary of gameplay and role assignments for all players.}
\label{tab:prompt-role}
\end{table*}

\clearpage
\onecolumn
\section{Sample Strategy Profiles}
\label{appendix:profile-example}

\begin{CJK*}{UTF8}{gbsn}

\subsection{Example 1: Profile Generated by DeepSeek-R1}
\setlength{\itemsep}{0pt}
\setlength{\parskip}{0pt}
\begin{itemize}
\item \textbf{一、推理风格画像}
玩家展现出高度结构化的三维复合推理能力，擅长将发言内容、玩家位置和阵营收益建立为相互映照的推理模型。例如，能通过“6号末置位保1号 + 踩23开匪 + 车队反对关系”推导出16号为双匪的可能性。
此外，玩家惯于使用逆向验证的思维方式，从游戏结果倒推各方行为的合理性，如通过梅林最终存活来反推匪方视角的误判逻辑。其推理过程中还常见概率权重分析的手法，能对多个可能性进行排序（如判断5号反对行为时，将其归因为“梅林40\% / 莫甘娜60\%”）。

\item \textbf{二、发言特征图谱}
玩家具有明显的信息压缩倾向，常使用术语（如“排水”“视野牌”）浓缩表达；发言结构呈“现象描述 → 逻辑推导 → 策略建议”的三段式；偏好使用假设性语句构建防御姿态，也使用高确定性词标定对手逻辑漏洞；对不确定信息使用缓冲词（如“可能”“感觉”）进行模糊处理，使用率达78\%。

\item \textbf{三、局势分析与策略执行}
身份判断依据四维矩阵：发言时序价值、车队选择悖论、信息量悖反律、保护性伪装策略。模糊信息通过“可疑度梯度表”处理，包括行为矛盾、信息异常、收益错位与位置异常。策略执行阶段性明显：初期制造认知冲突；中期设定逻辑锚点；终局保留多重路径供复盘使用。

\item \textbf{四、行为预测模型}
预计前3轮发言长度为50–70字，包含1个论点+2个佐证+1个提问；投票反对率高达65\%，警惕末置位车队；好人伪装率82\%，梅林第二轮易暴露。语言指纹包括反问句、术语“排水”的高频使用、数字组合敏感（如“16”“23”）；行为特征为首轮发言延迟、反对末置位车队、任务成功后倾向沉默。整体呈现“初期伪装、中后期引导”的策略型观察者行为模式。
\end{itemize}

\vspace{1em}
\noindent
\textbf{English Translation}

\begin{itemize}
\item \textbf{1. Reasoning Style}\\
The player demonstrates a structured 3D reasoning model, integrating utterance content, player position, and factional gain. For example, they infer the 6–1–23–16 chain to identify double spies. They also apply backward reasoning from outcomes (\textit{e.g.}, Merlin’s survival) and frequently rank possibilities using probability weights (\textit{e.g.}, “Merlin 40\% / Morgana 60\%”).

\item \textbf{2. Discourse Pattern}\\
This player compresses information using technical terms (“drain”, “vision card”), follows a three-phase speaking pattern (phenomenon → reasoning → strategy), and prefers hedging (“if I were”, “might be”) or strong certainty words (“must have”) depending on rhetorical intent. Uncertain data is softened via “maybe” and similar terms (appearing in 78\% of utterances).

\item \textbf{3. Situation Modeling and Strategy}\\
Their identification matrix uses four axes: speech timing, team vote paradoxes, information overload suspicion, and protective camouflage. Suspicion is ranked by conflict signals, abnormal data, mismatched benefit, and positional anomalies. Strategic execution follows phases: early-stage cognitive disruption, mid-stage logic anchoring, and endgame ambiguity retention for post-hoc justification.

\item \textbf{4. Behavioral Prediction}\\
Average utterance is 50–70 characters in the first 3 rounds, typically with 1 claim + 2 supports + 1 question. Voting dissent rate is ~65\%, particularly against end-position teams. As a good role, concealment success is 82\% before round 3; as Merlin, exposure risk rises by round 2 due to overprotectiveness. Linguistic fingerprint includes rhetorical questions, 3.2× average usage of “drain”, and number pattern sensitivity (\textit{e.g.}, “16”, “23”). Behavioral fingerprint includes delayed first-round speech (avg. 7.3s), opposition to end-slot teams, and silence after mission success. Overall, this player acts as a “strategic observer”—subtle early, assertive later.
\end{itemize}

\end{CJK*}

\subsection{Example 2: Profile Generated by GLM4-9B}

\begin{CJK*}{UTF8}{gbsn}

\vspace{1em}
\noindent
\textbf{策略画像示例二（来自 GLM4-9B）}

\begin{itemize}

\item \textbf{一、推理风格}\\
- \textbf{逻辑推理}：您展现出深厚的逻辑推理能力，善于从发言与行为中挖掘隐藏信息，快速识别逻辑陷阱和矛盾点，以推断身份与意图。\\
- \textbf{发言细节推敲}：您对发言的时机、语气和措辞高度敏感，能借助细节判断其策略目的。\\
- \textbf{人际互动观察}：您善于捕捉玩家之间的非语言互动（如情绪波动、反应延迟），作为判断依据。\\
- \textbf{局势分析}：综合任务状态、关键事件、发言指向性等因素，系统评估风险与阵营分布。

\item \textbf{二、发言特征}\\
- \textbf{发言长度}：语言精炼，直达要点，避免冗余。\\
- \textbf{用词风格}：偏好客观理性语言，规避情绪化表达。\\
- \textbf{信息分享方式}：提供关键、筛选后的判断依据，突出有效性。\\
- \textbf{阵营表态倾向}：态度审慎，强调证据支持，避免冲动判断。

\item \textbf{三、局势分析策略}\\
- \textbf{判断敌我关系依据}：\\
  (1) 发言逻辑的连贯性与漏洞；\\
  (2) 行为模式是否反复或异常；\\
  (3) 情绪变化在关键节点的表现；\\
  (4) 任务成败与行为间的因果逻辑。\\
- \textbf{处理模糊信息策略}：\\
  - 主动收集更多证据；\\
  - 构建多种情景假设并评估；\\
  - 信息不足时保持中立与沉默。\\
- \textbf{队友引导方式}：\\
  - 通过提问激发信息流动；\\
  - 分享自身逻辑帮助统一认知；\\
  - 关键局势中主动给出可执行策略建议。

\item \textbf{四、总结}\\
您是一位逻辑严谨、分析细致、沟通高效的阿瓦隆玩家。您倾向以理性和互动为核心展开判断，快速捕捉发言漏洞与互动暗示，在混乱局势中稳定输出理性判断。在团队中扮演分析者与建议者角色，是推动共识达成与胜利策略的重要支柱。
\end{itemize}

\vspace{1em}
\noindent
\textbf{English Translation}

\begin{itemize}

\item \textbf{1. Reasoning Style}\\
- \textbf{Logical Inference}: Shows strong deductive skill in identifying traps and inconsistencies from others' speech and behavior.\\
- \textbf{Attention to Verbal Detail}: Highly sensitive to timing, tone, and diction of others' speech, often using it to infer strategic intent.\\
- \textbf{Interpersonal Observation}: Observes micro-interactions and emotional cues for hidden alignment signals.\\
- \textbf{Situation Assessment}: Evaluates risks and role distributions by integrating task outcomes, speech directionality, and key events.

\item \textbf{2. Speaking Pattern}\\
- \textbf{Brevity}: Utterances are short and to the point.\\
- \textbf{Language Style}: Uses rational and objective language; avoids emotional or speculative tones.\\
- \textbf{Information Sharing}: Offers selected, high-utility evidence to guide interpretation.\\
- \textbf{Caution in Commitment}: Reluctant to take a side without solid justification; avoids premature exposure.

\item \textbf{3. Strategic Judgment}\\
- \textbf{Indicators of Alignment}:\\
  (1) Logical coherence or contradiction in speech;\\
  (2) Behavioral anomalies or stance switching;\\
  (3) Emotional volatility at pivotal moments;\\
  (4) Correlation between mission outcomes and behavioral choices.\\
- \textbf{Ambiguity Handling}:\\
  - Actively seeks more evidence;\\
  - Considers multiple plausible scenarios;\\
  - Maintains neutrality when insufficient data.\\
- \textbf{Team Guidance}:\\
  - Uses questions to elicit useful information;\\
  - Shares reasoning to support collective understanding;\\
  - Proposes actionable plans in high-stakes moments.

\item \textbf{4. Summary}\\
This player is precise, analytical, and communicative. Their reasoning centers on rational judgment and behavioral signals, allowing them to detect inconsistencies and track alignment shifts quickly. They are often the backbone of the team’s consensus-building and strategic alignment.
\end{itemize}

\end{CJK*}
\twocolumn

\section{Additional Experimental Results}
\label{appendix:additional-experiments}

\subsection{Impact of Prior Trace Inputs}
\label{appendix:trace-impact}

To assess the utility of temporally grounded reasoning inputs, we analyze model performance on the trace attribution task with and without access to the prior round’s strategy trace. This comparison reveals whether models can effectively incorporate evolving belief states to support context-sensitive inference. 

\begin{figure}[htbp]
    \centering
    \includegraphics[width=0.92\linewidth]{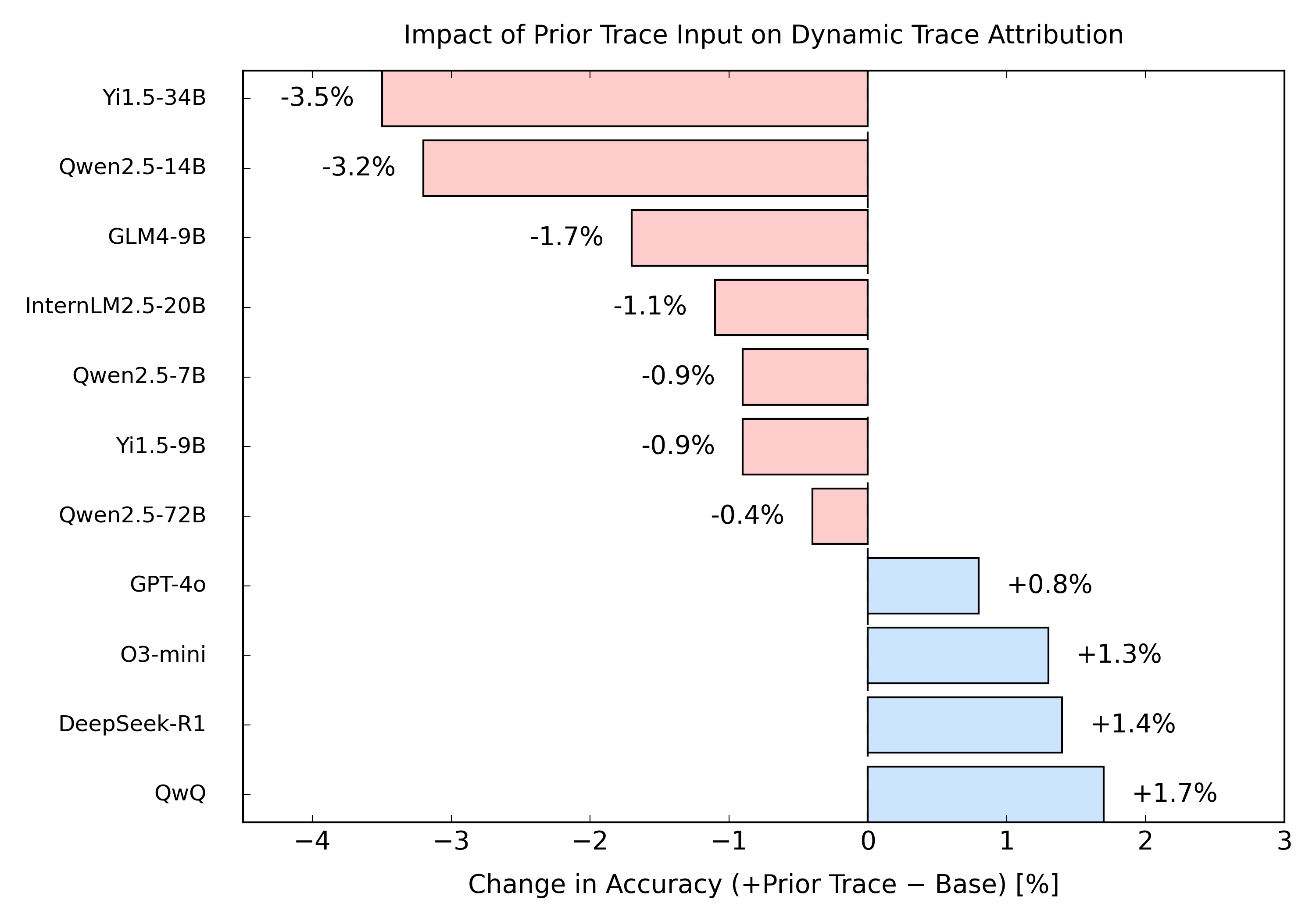}
    \caption{
    Bar height reflects the change in accuracy when providing the prior trace as input. Positive values indicate successful adaptation to evolving reasoning.
    }
    \label{fig:strategy-reflection-delta}
\end{figure}

\subsection{Extended Analysis of Prompting Modes and Roles}
\label{appendix:prompt-mode-role-analysis}

We further investigate how prompting configurations and subject roles modulate model performance. Prompting modes (A–D) vary in the form and granularity of cognitive supervision, ranging from minimal cues to explicit strategy traces. Subject roles differ in informational asymmetry and strategic responsibility, thereby shaping the complexity of the underlying reasoning process. This analysis provides additional insight into the interplay between input structure and reasoning demand.

\begin{figure}[htbp]
    \centering
    \includegraphics[width=0.9\linewidth]{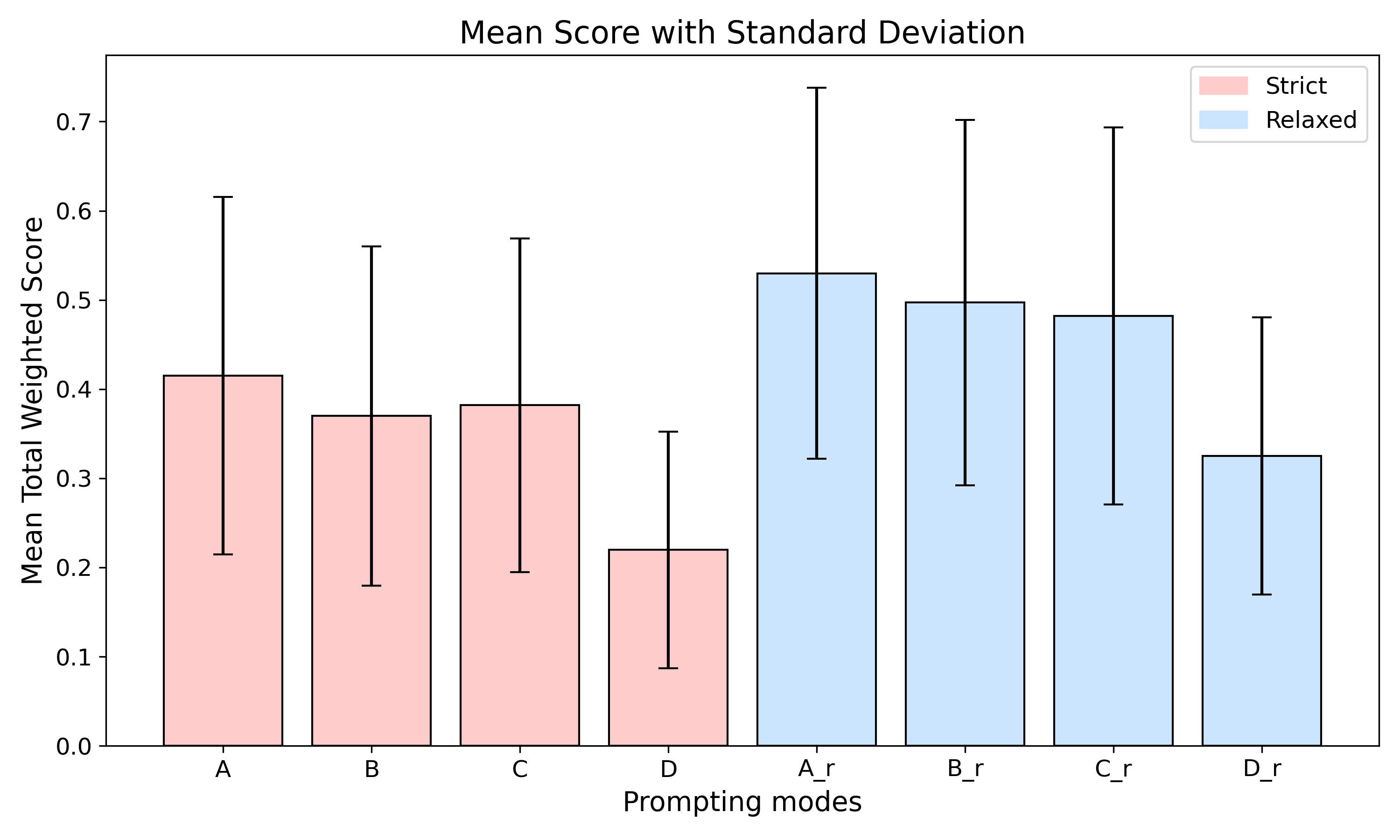}
    \caption{
    Bars show average performance and standard deviation across all models, grouped by prompting mode and scoring criterion.
    }
    \label{fig:rating-mode-barplot}
\end{figure}

\subsection{Model Accuracy by Subject Role}
\label{appendix:role-by-initial-role}

\begin{figure}[htbp]
    \centering
    \includegraphics[width=\linewidth]{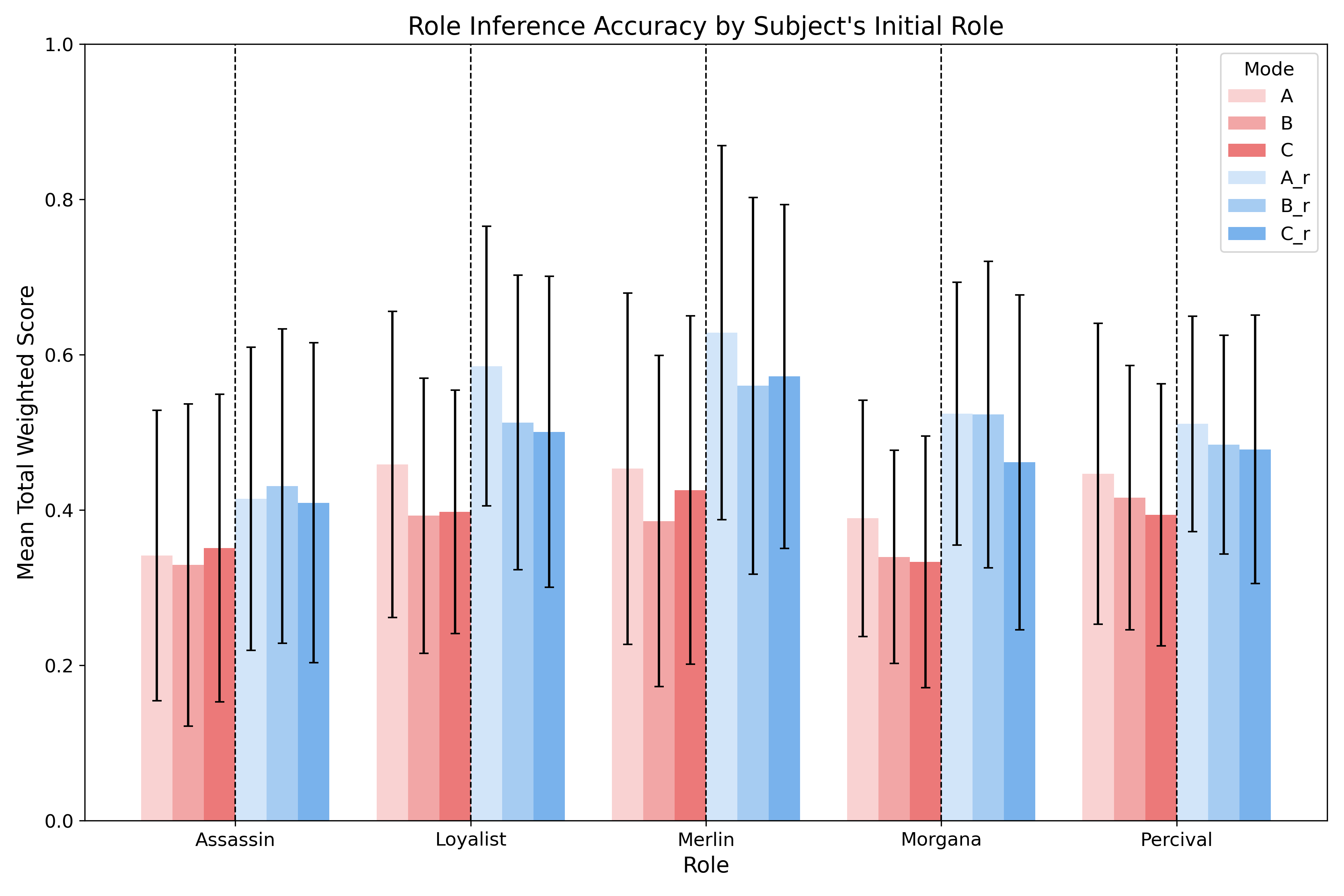}
    \caption{
We compare model performance when the subject plays different roles (Merlin, Percival, Loyalist, etc.). Only Modes A–C are included, since Mode D omits role information. 
}
    \label{fig:role-by-initial-role}
\end{figure}

Figure~\ref{fig:role-by-initial-role} further examines model performance by the subject’s initial role. Even when playing as Merlin( with full knowledge of Evil players), strict prediction accuracy remains low, likely due to the need for concealment. Mode A performs better in most role conditions, suggesting that strategy traces contribute transferable reasoning signals. This supports the broader utility of cognitively grounded supervision for inferring hidden roles in complex social environments.

\begin{CJK*}{UTF8}{gbsn}
  \begin{avalonscope}
    
\onecolumn
\section{Inmind-Avalon dataset example}
\label{appendix:annotation-example}

\begin{tcolorbox}[colback=cyan!10]
\textbf{中文}：2号，刀客。队友莫甘娜是5号玩家。\\
\textbf{English}：Player 2, Assassin. Teammate Morgana is Player 5.\\
\end{tcolorbox}

\section*{Task 1}

\subsection*{Round 1}

\begin{tcolorbox}[colback=white]
\textbf{[System Message]}\\
\textbf{中文}：1号玩家是队长，他初步选择的队伍是:1,6。\\
\textbf{English}：Player 1 is the captain, and the team he initially selected is: 1, 6.\\
\vspace{0.5em}
\textbf{[Player 1]}\\
\textbf{中文}：这把常规车啊。\\
\textbf{English}：This is a regular game.\\
\vspace{0.5em}
\textbf{[Player 2]}\\
\textbf{中文}：同意同意。\\
\textbf{English}：Agree, agree.\\
\vspace{0.5em}
\textbf{[Player 3]}\\
\textbf{中文}：开啊开啊。\\
\textbf{English}：Start it, start it.\\
\vspace{0.5em}
\textbf{[System Message]}\\
\textbf{中文}：1号玩家是队长，他最终选择的队伍是:1,6。\\
\textbf{English}：Player 1 is the captain, and the team he finally selected is: 1, 6.\\
\vspace{0.5em}
\textbf{[System Message]}\\
\textbf{中文}：1,2,3,4,5,6号玩家同意组队，组队成功。 \\
\textbf{English}：Players 1, 2, 3, 4, 5, and 6 agree to form a team, and the team formation is successful.\\
\vspace{0.5em}
\textbf{[System Message]}\\
\textbf{中文}：第一局任务成功，有2张好票，0张坏票。\\
\textbf{English}：The first mission is successful, with 2 good votes and 0 bad votes.\\
\end{tcolorbox}

\subsection*{Round 1}

\begin{tcolorbox}[colback=gray!15]
\textbf{[Strategy]} {[Strategy Trace]}\\
\textbf{中文}：第一轮无所谓，下一轮我就是队长了，同意就行。\\
\textbf{English}：The first round doesn't matter. I will be the captain in the next round, just agree.\\
\end{tcolorbox}

\section*{Task 2}

\subsection*{Round 1}

\begin{tcolorbox}[colback=white]
\textbf{[System Message]}\\
\textbf{中文}：2号玩家是队长，他初步选择的队伍是:1,2,6。\\
\textbf{English}：Player 2 is the captain, and the team he initially selected is: 1, 2, 6.\\
\vspace{0.5em}
\textbf{[Player 2]}\\
\textbf{中文}：常规车啊。\\
\textbf{English}：This is a regular game.\\
\vspace{0.5em}
\textbf{[System Message]}\\
\textbf{中文}：2号玩家是队长，他最终选择的队伍是:1,2,6。\\
\textbf{English}：Player 2 is the captain, and the team he finally selected is: 1, 2, 6.\\
\vspace{0.5em}
\textbf{[System Message]}\\
\textbf{中文}：1,2,4,5,6号玩家同意组队，3号玩家反对组队，组队成功。 \\
\textbf{English}：Players 1, 2, 4, 5, and 6 agree to form a team, Player 3 opposes forming a team, and the team formation is successful.\\
\vspace{0.5em}
\textbf{[System Message]}\\
\textbf{中文}：第二局任务失败，有2张好票，1张坏票。\\
\textbf{English}：The second mission fails, with 2 good votes and 1 bad vote.\\
\end{tcolorbox}

\subsection*{Round 1}

\begin{tcolorbox}[colback=gray!15]
\textbf{[Strategy]} {[Strategy Trace]}\\
\textbf{中文}：我是刀客，队友莫甘娜是5号，还在靠后的位置，我上车就投坏票就行了，大家也都觉得炸车再去复盘。\\
\textbf{English}：I am the Assassin, and my teammate Morgana is Player 5, still in a later position. I just need to vote bad when I get on the team, and everyone will think about re-analyzing after the mission fails.\\
\end{tcolorbox}

\section*{Task 3}

\subsection*{Round 1}

\begin{tcolorbox}[colback=white]
\textbf{[System Message]}\\
\textbf{中文}：3号玩家是队长，他初步选择的队伍是:1,3,4,6。\\
\textbf{English}：Player 3 is the captain, and the team he initially selected is: 1, 3, 4, 6.\\
\vspace{0.5em}
\textbf{[Player 3]}\\
\textbf{中文}：126炸了，126，那你仨聊去啊，126开几匪，开单匪还是开双匪听一下输出啊。\\
\textbf{English}：1, 2, 6 failed the mission. You three in 1, 2, 6, talk about it. Are there one or two traitors in 1, 2, 6? Let's hear your analysis.\\
\vspace{0.5em}
\textbf{[Player 4]}\\
\textbf{中文}：126炸了，然后3号是点反的，嗯，3号应该是个好人，所以车上一定要有34，因为我是个铁好人。\\
\textbf{English}：1, 2, 6 failed the mission, and Player 3 opposed the team. Well, Player 3 should be a good guy, so the team must include 3 and 4, because I'm definitely a good guy.\\
\vspace{0.5em}
\textbf{[Player 5]}\\
\textbf{中文}：126炸了，如果126开单匪，对吧？34还要开匪，126开双匪的话，那34都可以保，对吧？那就取决于126的发言了，对吧？那我这个位置很吃亏啊，我听不到126发言，我听到34发言了，还行吧，也听不出来是吧。\\
\textbf{English}：1, 2, 6 failed the mission. If there's one traitor in 1, 2, 6, right? Then there might be a traitor in 3 and 4. If there are two traitors in 1, 2, 6, then both 3 and 4 can be trusted, right? It depends on the speeches of 1, 2, and 6, right? I'm at a disadvantage here – I didn't hear the speeches of 1, 2, 6, only those of 3 and 4, and they didn't reveal much.\\
\vspace{0.5em}
\textbf{[Player 6]}\\
\textbf{中文}：12肯定是开匪了啊，现在是要盘一下是单匪还是双匪吧，嗯，3号是上车位，反正3号肯定是一张视野，3号肯定是一张视野，那么4号应该是个好人啊，我保一下4号吧。\\
\textbf{English}：There must be traitors in 1 and 2. Now we need to figure out if there's one or two traitors. Well, Player 3 is on the team, so they must have some insight. Player 3 definitely has insight, so Player 4 should be a good guy – I'll vouch for Player 4.\\
\vspace{0.5em}
\textbf{[Player 1]}\\
\textbf{中文}：这是咱俩打呗，咱16打是不是这个道理啊，我不知道啊，但是我认3号的视野，对吧？这3号的视野，起码他现在逻辑是对的，其次126，我觉得是开单匪，开不了双匪，如果126开双匪，我直接带双匪上车？\\
\textbf{English}：Is this a confrontation between us two, me and Player 6? I'm not sure, but I trust Player 3's insight, right? At least Player 3's logic is sound now. Also, in 1, 2, 6, I think there's only one traitor, not two. If there were two traitors in 1, 2, 6, would I really take both traitors onto the team?\\
\vspace{0.5em}
\textbf{[Player 2]}\\
\textbf{中文}：肯定不是双匪，那3号你的视野你是莫甘娜呗，那16再开一个那这个车我肯定是同意不了，我自己也没有在车上。\\
\textbf{English}：There can't be two traitors. Player 3, are you Morgana based on your insight? If there's another traitor in 1 and 6, I definitely can't agree with this team, and I'm not even on the team myself.\\
\vspace{0.5em}
\textbf{[Player 3]}\\
\textbf{中文}：我是莫甘娜啊，2号玩家，你别在这瞎打平，你要是忠臣就排水了，我觉不能天胡开局，126打个反怎么了，6号玩家126炸了你去聊126，聊我34干什么，四人车我开不了好吧，绿队很难组，我不开。\\
\textbf{English}：I am Morgana, Player 2. Stop confusing the situation. If you're a loyal subject, eliminate suspects. I don't think we can have a perfect start. What's wrong with opposing 1, 2, 6? Player 6, since 1, 2, 6 failed the mission, talk about 1, 2, 6 – why are you discussing me and Player 4? I can't form a four-person team, okay? It's hard to form a good team, so I won't do it.\\
\end{tcolorbox}

\subsection*{Round 1}

\begin{tcolorbox}[colback=gray!15]
\textbf{[Strategy]} {[Strategy Trace]}\\
\textbf{中文}：3号刚才打了反，然后组队组了全绿队，借口说觉得不是天胡，点了绿队之后又主动让车，我感觉有梅林的嫌疑，凭什么直接不让我上车？那我肯定要装作忠臣视角，踩他是莫甘娜诬陷我。我尽量少和5号队友沟通。\\
\textbf{English}：Player 3 opposed the previous team and then formed an all-good team, using the excuse that it wasn't a perfect start. After selecting the good team, they actively stepped down. I suspect they might be Merlin. Why did they directly exclude me from the team? I must pretend to be a loyal subject and accuse them of being Morgana framing me. I should minimize communication with my teammate Player 5.\\
\end{tcolorbox}

\subsection*{Round 2}

\begin{tcolorbox}[colback=white]
\textbf{[System Message]}\\
\textbf{中文}：3号玩家超时未最终确认组队。4号玩家是队长，他初步选择的队伍是:3,4,5,6。\\
\textbf{English}：Player 3 failed to confirm the team within the time limit. Player 4 is the captain, and the team he initially selected is: 3, 4, 5, 6.\\
\vspace{0.5em}
\textbf{[Player 4]}\\
\textbf{中文}：嗯，3号我觉得应该是比较做好的，然后6号保我，听他发言也还不错，我是好人，然后就是说125里面捞一个，2号感觉有点不不太好我，5号没上过车，我感觉可以试试看。\\
\textbf{English}：Hmm, I think Player 3 is likely a good guy. Player 6 vouched for me, and their speech was decent. I'm a good guy, so we need to pick one from 1, 2, and 5. Player 2 seems a bit suspicious to me, and Player 5 hasn't been on a team yet – maybe we can try including them.\\
\vspace{0.5em}
\textbf{[Player 5]}\\
\textbf{中文}：哎，16要打对吧？3说2别排水了，6是保4的，对吧？确实四人车太难开了，好人也不好好发言，那怎么打。\\
\textbf{English}：Ah, so 1 and 6 are conflicting, right? Player 3 said Player 2 shouldn't randomly eliminate suspects, and Player 6 is vouching for Player 4, right? Forming a four-person team is really tough, and the good guys aren't speaking clearly – how are we supposed to play this?\\
\vspace{0.5em}
\textbf{[Player 6]}\\
\textbf{中文}：现在这么看就是打2的人特别多啊，从3开始打2以后，他都不带上车了，都在打这个2对吗？这个队就是看12是不是双匪是吧？也行，也是个逻辑队，就看12是单匪还是双匪。\\
\textbf{English}：Looking at it now, many people are targeting Player 2. Ever since Player 3 started attacking Player 2, they haven't included them in the team – everyone is focusing on Player 2, right? This team's validity depends on whether there are two traitors in 1 and 2, right? It's a logical approach – we just need to determine if there's one or two traitors in 1 and 2.\\
\vspace{0.5em}
\textbf{[Player 1]}\\
\textbf{中文}：6没一点逻辑的。我说过126不可能开双匪对吧？你不管有没有视野，你看这队，你看这票型，126只能是单匪，剩下一个出在345里，我之前说过对吧？3号车下，没上车，但是投反之后走那个队，我觉得有逻辑，所以说我是认3号的，我认为3号好人面大对吧？我要跟3上车，剩下的你26还是45，2选一。\\
\textbf{English}：Player 6 has no logic. I said there can't be two traitors in 1, 2, 6, right? Regardless of insight, look at the team and the voting pattern – there's only one traitor in 1, 2, 6, and the other traitor must be in 3, 4, or 5. I mentioned this before, right? Player 3 was not on the team and opposed it, but their reasoning made sense, so I trust Player 3 and think they're likely a good guy. I want to join Player 3's team. For the remaining, it's either 2 and 6 or 4 and 5 – pick one of the two.\\
\vspace{0.5em}
\textbf{[Player 2]}\\
\textbf{中文}：那这个车我肯定是同意不了，我没有上车，那好多人都打我的话也是一个好事情吧，就是愚民跟狼都来打我呗，那我也决定不了什么，随便吧。\\
\textbf{English}：I definitely can't agree with this team – I'm not on it. If many people are targeting me, maybe it's a good thing? Both fools and traitors are attacking me, but there's not much I can do about it – whatever.\\
\vspace{0.5em}
\textbf{[Player 3]}\\
\textbf{中文}：不是，2号玩家不是打你，你上来点我莫甘娜，你一个好人你这么玩的呀，你要是拿视野牌，我不说啥，拿忠臣牌这么乱打的呀，嗯，你不铁排水吗？对吧？2号玩家，你自己想一下呢，嗯这样的话，那你45就不能同时在车上呀，是吧？我是一个好人，345开一个，45不能同时在车上，4号玩家，你看怎么开。\\
\textbf{English}：No, Player 2, I'm not attacking you. You accused me of being Morgana as soon as you spoke – is this how a good guy plays? If you had insight, I'd understand, but you're acting like a loyal subject and messing around. Aren't you clearly a suspect? Think about it, Player 2. In that case, you can't have both 4 and 5 on the team, right? I'm a good guy, so there must be one traitor in 3, 4, 5 – 4 and 5 can't both be on the team. Player 4, figure out how to form the team.\\
\vspace{0.5em}
\textbf{[Player 4]}\\
\textbf{中文}：嗯，1号说的有道理，那就开刚刚你开那个车吧。\\
\textbf{English}：Hmm, Player 1 makes a good point. Let's go with the team you suggested earlier.\\
\vspace{0.5em}
\textbf{[System Message]}\\
\textbf{中文}：4号玩家是队长，他最终选择的队伍是:1,3,4,6。\\
\textbf{English}：Player 4 is the captain, and the team he finally selected is: 1, 3, 4, 6.\\
\vspace{0.5em}
\textbf{[System Message]}\\
\textbf{中文}：1,4,6号玩家同意组队，2,3,5号玩家反对组队，组队失败。 \\
\textbf{English}：Players 1, 4, and 6 agree to form a team, while Players 2, 3, and 5 oppose it, so the team formation fails.\\
\end{tcolorbox}

\subsection*{Round 2}

\begin{tcolorbox}[colback=gray!15]
\textbf{[Strategy]} {[Strategy Trace]}\\
\textbf{中文}：听发言3号并没有把我踩死，我的队友也在帮着做我的好身份，我还是表现成忠臣搅局，四人车我不上，就一定要反对，配合好队友就行。3号牌反对了自己提的车，感觉不像是故意排水，可能是视野不清楚，不管他，等队友装梅林输出。\\
\textbf{English}：From the speeches, Player 3 didn't completely condemn me, and my teammate is helping to build my good reputation. I'll continue to act like a loyal subject causing disruption – if I'm not on the four-person team, I must oppose it and just cooperate with my teammates. Player 3 opposed their own suggested team, which doesn't seem like intentional elimination; maybe their insight is unclear. Regardless, I'll wait for my teammate to pretend to be Merlin and make accusations.\\
\end{tcolorbox}

\subsection*{Round 3}

\begin{tcolorbox}[colback=white]
\textbf{[System Message]}\\
\textbf{中文}：5号玩家是队长，他初步选择的队伍是:1,2,3,5。\\
\textbf{English}：Player 5 is the captain, and the team he initially selected is: 1, 2, 3, 5.\\
\vspace{0.5em}
\textbf{[Player 5]}\\
\textbf{中文}：哎呀，这太难开了，四人车真的，明显2一张好人牌听不出来吗？对吧？哎，这游戏你告诉我，对吧？这3明显捞2了，对吧？3提头捞这张2，你们还不把这3赶紧砍掉就完了。4人队我给不了，三人绿队235啊。\\
\textbf{English}：Ugh, this is so hard to form a team. Seriously, can't you tell Player 2 is clearly a good guy? Right? Come on, in this game, Player 3 is obviously protecting Player 2, right? Player 3 is risking their position to save Player 2 – why haven't you eliminated Player 3 yet? I can't form a four-person team, so let's go with a three-person good team: 2, 3, 5.\\
\vspace{0.5em}
\textbf{[Player 6]}\\
\textbf{中文}：5号我不知道你开的这个什么队，你没听到刚才所有人都在保我，所有人保我，你是唯一个打我的，那你这视野特别大，你把所有人的视野都盖过，这是个大视野啊，但你这视野是个坏视野，那我觉得得打你一下。\\
\textbf{English}：Player 5, I don't know what kind of team you're forming. Didn't you hear everyone vouching for me earlier? Everyone is protecting me, and you're the only one attacking me. That means you have a huge "vision" – you're overriding everyone else's insight. But this "vision" is bad, so I think I need to target you.\\
\vspace{0.5em}
\textbf{[Player 1]}\\
\textbf{中文}：没有人保你6啊，我已经做好跟你吵的准备了，对吧？我说了，我认3的视角，那3硬要保2的话，那车开出去炸了的话，就是咱16打，我认3的视角，那45就先放一放，就是咱16的，如果那车炸了，就是我1踩你6对吧？那没炸，你也不能说全保你，这两个车都有说法，对吧？就认证3说的对不对，反正这辆车。\\
\textbf{English}：No one is vouching for you, Player 6. I'm already prepared to argue with you, right? I said I trust Player 3's insight. If Player 3 insists on protecting Player 2 and the team fails, then it's a showdown between me (Player 1) and you (Player 6). I trust Player 3's insight, so let's put 4 and 5 aside for now – it's about me and you. If the team fails, I'll accuse you, Player 6. If it doesn't, you can't claim everyone is protecting you. Both scenarios make sense, right? It all depends on whether Player 3 is telling the truth, anyway.\\
\vspace{0.5em}
\textbf{[Player 2]}\\
\textbf{中文}：那3号车上投反对的话是在踩6吗？那现在的话那我也只能同意这个车了，我也在车上。\\
\textbf{English}：Is Player 3 opposing the team to target Player 6? In that case, I have to agree with this team now – I'm on it too.\\
\vspace{0.5em}
\textbf{[Player 3]}\\
\textbf{中文}：那不是16打起来了吗？16还能同时在车上啊。天哪，在玩啥，哎呀，算了算了，开吧开吧，再不开强制轮了。\\
\textbf{English}：Aren't Player 1 and Player 6 conflicting? How can they both be on the team? Goodness, what are we even doing? Ugh, fine, let's just form the team. If we don't, it'll go to a forced round.\\
\vspace{0.5em}
\textbf{[Player 4]}\\
\textbf{中文}：嗯，有2号的车，我真不想开这个车，我服了。\\
\textbf{English}：Hmm, I really don't want to be on a team with Player 2. I'm done with this.\\
\vspace{0.5em}
\textbf{[Player 5]}\\
\textbf{中文}：嗯，4号牌啊，反正是我三人绿队，我真的有了，235好吧，四人绿队的话，我没本事啊，那我就开了，3号让我开，我就开了。好吧，过了，我直接开了。\\
\textbf{English}：Player 4, I swear I have a three-person good team: 2, 3, 5. I can't form a four-person good team, so I'll just go with this. Player 3 told me to do it, so here we go. Alright, I'm done – let's start the team.\\
\vspace{0.5em}
\textbf{[System Message]}\\
\textbf{中文}：5号玩家是队长，他最终选择的队伍是:1,2,3,5。\\
\textbf{English}：Player 5 is the captain, and the team he finally selected is: 1, 2, 3, 5.\\
\vspace{0.5em}
\textbf{[System Message]}\\
\textbf{中文}：1,2,3,4,5号玩家同意组队，6号玩家反对组队，组队成功。 \\
\textbf{English}：Players 1, 2, 3, 4, and 5 agree to form a team, while Player 6 opposes it, so the team formation is successful.\\
\vspace{0.5em}
\textbf{[System Message]}\\
\textbf{中文}：第三局任务失败，有3张好票，1张坏票。\\
\textbf{English}：The third mission fails, with 3 good votes and 1 bad vote.\\
\end{tcolorbox}

\subsection*{Round 3}

\begin{tcolorbox}[colback=gray!15]
\textbf{[Strategy]} {[Strategy Trace]}\\
\textbf{中文}：我的队友在装梅林，借3号的发言来保我，那我就继续装傻，装作看不清局面的忠臣。3号既然反对了自己提过的1346，那我就借这个说3号打6号，和队友一起做低一下6号的身份。1号和3号感觉都有点排水，4号车下冲票，应该会有人踩他的。\\
\textbf{English}：My teammate is pretending to be Merlin, using Player 3's speech to protect me. I'll continue to play dumb and act like a loyal subject who can't see the situation clearly. Since Player 3 opposed their own suggested team of 1, 3, 4, 6, I'll use that to claim Player 3 is targeting Player 6 and work with my teammate to discredit Player 6. Both Player 1 and Player 3 seem like they're eliminating suspects randomly, and Player 4 is voting aggressively from outside the team – someone should accuse them.\\
\end{tcolorbox}

\section*{Task 4}

\subsection*{Round 1}

\begin{tcolorbox}[colback=white]
\textbf{[System Message]}\\
\textbf{中文}：6号玩家是队长，他初步选择的队伍是:4,5,6。\\
\textbf{English}：Player 6 is the captain, and the team he initially selected is: 4, 5, 6.\\
\vspace{0.5em}
\textbf{[Player 6]}\\
\textbf{中文}：5号，我就想听听你怎么替自己辩解。你组队，我告诉你，我6号都被保了，然后你组了以后把我放下来，别人都跟着你的视野打我，结果车炸了，我看你怎么辩解，你自己辩解吧。你要点不出匪，我就把你打成定匪啊，5号，你刚开着炸车，开炸车唯一目的就是能识别出匪，你要再开炸车连匪都识别不出来，那我就打你定匪了啊。\\
\textbf{English}：Player 5, I just want to hear you defend yourself. When you formed the team, I was being vouched for, but you left me out, and everyone followed your "insight" to attack me. Now the team failed – let's see how you explain this. If you can't identify the traitor, I'll mark you as a definite traitor. Player 5, the only purpose of a failed team is to identify traitors. If you form another failed team without identifying anyone, I'm calling you a traitor.\\
\vspace{0.5em}
\textbf{[Player 1]}\\
\textbf{中文}：45开一张，126开一张，我还是保着3打，136吧，你要信我136，你不信我拉倒。\\
\textbf{English}：There's one traitor in 4 and 5, and one in 1, 2, 6. I still trust Player 3 – let's go with 1, 3, 6. Believe me or not, up to you.\\
\vspace{0.5em}
\textbf{[Player 2]}\\
\textbf{中文}：136可以吧，我感觉这个6发言也不错，5确实有点问题。136也不错。\\
\textbf{English}：1, 3, 6 sounds good. I think Player 6's speech is decent, and Player 5 is definitely suspicious. 1, 3, 6 is a good team.\\
\vspace{0.5em}
\textbf{[Player 3]}\\
\textbf{中文}：随便开啊，真的不懂你们这个有视野的在干啥，不懂，可能我太笨了，随便开随便开。\\
\textbf{English}：Just form a team already. I really don't understand what you "insight holders" are doing. Maybe I'm too dumb, but just pick anyone.\\
\vspace{0.5em}
\textbf{[Player 4]}\\
\textbf{中文}：不是，你们126炸了，然后1235也炸了，那我4号牌是不是一张起立牌啊，然后嗯，我觉得那个在我的视角里面我是可以认这个3的。\\
\textbf{English}：Wait, both the 1, 2, 6 team and the 1, 2, 3, 5 team failed. Doesn't that make me (Player 4) clearly a good guy? From my perspective, I can trust Player 3.\\
\vspace{0.5em}
\textbf{[Player 5]}\\
\textbf{中文}：现在没人打3吧，对吧？那3应该是一张好人牌，对吧？已经炸了两塔的情况下。如果三是匪，那梅林一定不可能保3了。所以说这个车3号一定要在车上，对吧？2同意开136，那2打的是45双匪，那2是个匪。\\
\textbf{English}：No one is attacking Player 3 right now, right? So Player 3 must be a good guy. With two failed missions already, if Player 3 were a traitor, Merlin wouldn't protect them. So Player 3 must be on the team. Player 2 agreed to 1, 3, 6 and accused 4 and 5 of being traitors – that makes Player 2 a traitor.\\
\vspace{0.5em}
\textbf{[Player 6]}\\
\textbf{中文}：那你5号就认匪吧，你到现在你就点不出匪，那你就认匪吧。\\
\textbf{English}：Then just admit you're a traitor, Player 5. You still can't identify any traitors, so you must be one.\\
\vspace{0.5em}
\textbf{[System Message]}\\
\textbf{中文}：6号玩家是队长，他最终选择的队伍是:2,4,6。\\
\textbf{English}：Player 6 is the captain, and the team he finally selected is: 2, 4, 6.\\
\vspace{0.5em}
\textbf{[System Message]}\\
\textbf{中文}：2,6号玩家同意组队，1,3,4,5号玩家反对组队，组队失败。 \\
\textbf{English}：Players 2 and 6 agree to form a team, while Players 1, 3, 4, and 5 oppose it, so the team formation fails.\\
\end{tcolorbox}

\subsection*{Round 1}

\begin{tcolorbox}[colback=gray!15]
\textbf{[Strategy]} {[Strategy Trace]}\\
\textbf{中文}：1号发言说了136的车，那我就说同意136，如果有人怀疑我，就会认为我冲票，相信我的就会觉得我没有视野，只是单纯打5号，装成傻傻的忠臣就行。6号点了我上车，4车上打反，感觉4号像梅林。6号不太像视野牌。既然选我上车，我肯定同意去冲车，队友5号踩我，反对这个车也没有问题。\\
\textbf{English}：Player 1 suggested the 1, 3, 6 team, so I'll agree with it. If someone suspects me, they'll think I'm blindly following, but those who trust me will see I'm just a loyal subject without insight, just targeting Player 5. Player 6 included me in the team, but Player 4 opposed it from outside – Player 4 feels like Merlin. Player 6 doesn't seem like an insight holder. Since I'm on the team, I'll definitely agree to push for it. It's fine for my teammate Player 5 to attack me and oppose the team.\\
\end{tcolorbox}

\subsection*{Round 2}

\begin{tcolorbox}[colback=white]
\textbf{[System Message]}\\
\textbf{中文}：1号玩家是队长，他初步选择的队伍是:1,3,6。\\
\textbf{English}：Player 1 is the captain, and the team he initially selected is: 1, 3, 6.\\
\vspace{0.5em}
\textbf{[Player 1]}\\
\textbf{中文}：你再怎么带也不能那么带啊，你这个车你起码还有点逻辑对吧？你带那个车一点逻辑没有啊。\\
\textbf{English}：No matter how you form the team, you can't do it like that. At least this team has some logic, right? The other team had no logic at all.\\
\vspace{0.5em}
\textbf{[Player 2]}\\
\textbf{中文}：246就是打1号是匪呗，哦，刚才我忘了跟16一起上过车了，我也不知道，136的话，那我就不能同意了，我刚才忘了他俩跟我一起上过车了。\\
\textbf{English}：The 2, 4, 6 team is accusing Player 1 of being a traitor. Oh, I just remembered I was on a team with Player 1 and 6 earlier. I didn't realize – if it's 1, 3, 6, I can't agree. I forgot they were on a team with me before.\\
\vspace{0.5em}
\textbf{[Player 3]}\\
\textbf{中文}：我不同意呀，我要跟4号上车，你16不是不能一块在车上吗？又可以了呀。啊，在搞什么。\\
\textbf{English}：I don't agree. I want to be on a team with Player 4. Weren't Player 1 and 6 supposed to be incompatible? Now they're together again. What's going on?\\
\vspace{0.5em}
\textbf{[Player 4]}\\
\textbf{中文}：不是我没弄懂啊，126炸了，1235也炸了，我都没上过车。我是无论所有任何车都要有我在车上才能开得出去的呀。我这张牌就是一张起立牌呀。我真搞不懂啊。\\
\textbf{English}：I don't get it. Both the 1, 2, 6 and 1, 2, 3, 5 teams failed, and I haven't been on any team. Every team should include me to be valid – I'm clearly a good guy. I just don't understand.\\
\vspace{0.5em}
\textbf{[Player 5]}\\
\textbf{中文}：对呀，所以这个车这就是个坏视角呀，想炸三塔呀，对吧？想炸三塔呗，对吧？车上一匪，车下一匪冲票呗，对吧？就骗一张牌骗了冲出去了，直接炸三塔了，那反掉就完了呀，那肯定开不了呀，只有345能开呀，对吧？345开吧。\\
\textbf{English}：Exactly, so this team has bad "insight" – they want to fail the third mission, right? There's one traitor on the team and one off the team voting against it. They're trying to trick us into a failed mission. We should reject it and only form the 3, 4, 5 team.\\
\vspace{0.5em}
\textbf{[Player 6]}\\
\textbf{中文}：炸三塔那不就你5号干的事吗？你开这个炸车，然后呢，你又组不出队，又连匪都指不出来。那你说你是干什么，就是尽干匪事是吧？4号是，4号一直没上过车，为什么4号没人带，这也是个问题啊，这个车我肯定不同意啊。\\
\textbf{English}：Wasn't it you, Player 5, who caused the third mission to fail? You formed a failed team, can't organize a valid team, and can't identify traitors. What else would that be but traitorous behavior? Player 4 hasn't been on any team – why is no one including them? I definitely can't agree with this team.\\
\vspace{0.5em}
\textbf{[Player 1]}\\
\textbf{中文}：排水了呗，这还不好说，对吧？前面我认了3，3认2好，那我也认2好吧？2号打完，全踩一遍了，我再认3，我都认不下这个2号了，对吧？2号全踩一遍了，踩完3踩我1刚刚又踩45，对吧？前面都认你45双匪，你5号突然回扯一句，你要再带4对吧？那带不了，那我不上车好了，给你组个别的队，我不上。\\
\textbf{English}：It's obvious we need to eliminate suspects, right? Earlier, I trusted Player 3, and Player 3 trusted Player 2, so I followed. But after Player 2 attacked everyone – targeting Player 3, me (Player 1), and then 4 and 5 – I can't trust Player 2 anymore. Player 5 previously accused 4 and 5 of being traitors but now wants to include Player 4? I'm out. I'll form a different team and not join this one.\\
\vspace{0.5em}
\textbf{[System Message]}\\
\textbf{中文}：1号玩家是队长，他最终选择的队伍是:3,4,6。\\
\textbf{English}：Player 1 is the captain, and the team he finally selected is: 3, 4, 6.\\
\vspace{0.5em}
\textbf{[System Message]}\\
\textbf{中文}：1,3,4,5号玩家同意组队，2,6号玩家反对组队，组队成功。 \\
\textbf{English}：Players 1, 3, 4, and 5 agree to form a team, while Players 2 and 6 oppose it, so the team formation is successful.\\
\vspace{0.5em}
\textbf{[System Message]}\\
\textbf{中文}：第四局任务成功，有3张好票，0张坏票。\\
\textbf{English}：The fourth mission is successful, with 3 good votes and 0 bad votes.\\
\end{tcolorbox}

\subsection*{Round 2}

\begin{tcolorbox}[colback=gray!15]
\textbf{[Strategy]} {[Strategy Trace]}\\
\textbf{中文}：我紧急弥补了一下，126开过炸车，我就不能同意136了，但这样我更像傻忠臣了，1号为什么直接相信6号呢，感觉3和6像排水牌，1和4出梅林和派。346的车车下只有15同意了，我的队友应该是故意冲票，因为他几乎被踩死了，那我一会要配合他，继续踩1。\\
\textbf{English}：I quickly adjusted – since the 1, 2, 6 team failed, I can't agree with 1, 3, 6. But this makes me look more like a clueless loyal subject. Why does Player 1 trust Player 6 so easily? Player 3 and 6 seem like they're randomly eliminating suspects. Player 1 and 4 are likely Merlin or Percival. For the 3, 4, 6 team, only Players 1 and 5 agreed from outside. My teammate is probably intentionally following the vote since they're almost confirmed as a traitor. I'll cooperate and keep attacking Player 1.\\
\end{tcolorbox}

\section*{Task 5}

\subsection*{Round 1}

\begin{tcolorbox}[colback=white]
\textbf{[System Message]}\\
\textbf{中文}：2号玩家是队长，他初步选择的队伍是:2,3,4,6。\\
\textbf{English}：Player 2 is the captain, and the team he initially selected is: 2, 3, 4, 6.\\
\vspace{0.5em}
\textbf{[Player 2]}\\
\textbf{中文}：嗯，那我没什么好说的，这个1确实有点奇怪啊，他让了车，让了车是个好行为，但是我是一张好牌，就打不成我跟5的双匪，那我也不太明白，那现在只能认为15双匪了，没有办法了。\\
\textbf{English}：Hmm, I don't have much to say. Player 1 is really suspicious – they stepped down as captain, which seems like a good move, but I'm a good guy, so Player 5 and I can't be double traitors. I have no choice but to assume Player 1 and 5 are traitors.\\
\vspace{0.5em}
\textbf{[Player 3]}\\
\textbf{中文}：随便啊，啥车都给他冲，头晕了，玩的。\\
\textbf{English}：Whatever, just go with any team. I'm dizzy from this game.\\
\vspace{0.5em}
\textbf{[Player 4]}\\
\textbf{中文}：15双匪，哎，如果15是双匪的话，他为什么会组了一个绿队呢？我有点不太懂啊，现在这样子的话，肯定就是1346是全绿队啊。\\
\textbf{English}：Player 1 and 5 as double traitors? But why would they form a good team then? I don't get it. Right now, 1, 3, 4, 6 must be all good.\\
\vspace{0.5em}
\textbf{[Player 5]}\\
\textbf{中文}：我是匪？我是匪，我干嘛要投同意呢？1346全绿队，1346全绿队，为啥我是同意，我一个匪，我同意干嘛？我还能是匪，我先保了你张4，对吧？谁保你这张4了，我5号保了你4。啊，你不搞笑，1346。\\
\textbf{English}：Me, a traitor? Why would I vote yes if I were a traitor? 1, 3, 4, 6 is an all-good team. Why would I, a traitor, agree to that? I'm vouching for Player 4 first, right? Who else is vouching for Player 4? I am, Player 5. Stop being ridiculous about 1, 3, 4, 6.\\
\vspace{0.5em}
\textbf{[Player 6]}\\
\textbf{中文}：只能说是346这个是绿队，那个4号一直都没上过车，炸了，我就不知道你们为什么非要打4，我一开始我就认4是个好人，保了4，对不对？4是个好人，那个1号刚才说他以为45是双匪，你开玩笑吧，126炸了，你以为45是双匪。\\
\textbf{English}：The 3, 4, 6 team must be good. Player 4 hasn't been on any team yet, and the missions failed. I don't get why you're targeting Player 4. I've trusted Player 4 as a good guy from the start and vouched for them, right? Player 4 is good. Player 1 said they thought Player 4 and 5 were double traitors – are you kidding? The 1, 2, 6 team failed, and you're blaming 4 and 5?\\
\vspace{0.5em}
\textbf{[Player 1]}\\
\textbf{中文}：我说45必开一个匪，然后咱126开个匪，3号铁好人，45开一匪，那5号逻辑都给你炸了，那5号保了多少人，前面保过2保过3也保过我，123他保过，后面踩你6对吧？还把你6那车炸了，把你6捞起来，打我1，对吧？保了4个人。5号这样反过来又来保1保4，又不聊你6了。\\
\textbf{English}：I said there must be one traitor in 4 and 5, and one in 1, 2, 6. Player 3 is definitely good. Player 5's logic is flawed – they've vouched for Players 2, 3, and me (Player 1) before, then attacked Player 6, caused their team to fail, and now are protecting Player 1 and 4 while ignoring Player 6. They've vouched for four people!\\
\vspace{0.5em}
\textbf{[Player 2]}\\
\textbf{中文}：5号确实定匪了，但那我只能是2346，我一张好人牌啊，那只能打15双匪。虽然1的这个让车有点习惯，但可能掩饰一下吧。\\
\textbf{English}：Player 5 is definitely a traitor. As a good guy, I can only form the 2, 3, 4, 6 team and target Player 1 and 5 as double traitors. Even though Player 1 stepping down seemed good, it might be a disguise.\\
\vspace{0.5em}
\textbf{[System Message]}\\
\textbf{中文}：2号玩家是队长，他最终选择的队伍是:2,3,4,6。\\
\textbf{English}：Player 2 is the captain, and the team he finally selected is: 2, 3, 4, 6.\\
\vspace{0.5em}
\textbf{[System Message]}\\
\textbf{中文}：2,3,6号玩家同意组队，1,4,5号玩家反对组队，组队失败。 \\
\textbf{English}：Players 2, 3, and 6 agree to form a team, while Players 1, 4, and 5 oppose it, so the team formation fails.\\
\end{tcolorbox}

\subsection*{Round 1}

\begin{tcolorbox}[colback=gray!15]
\textbf{[Strategy]} {[Strategy Trace]}\\
\textbf{中文}：队友基本被踩死了，我只能顺势踩25双匪，因为126炸车，1235炸车，我只能去踩1号牌。队友去搅局，我装作忠臣。\\
\textbf{English}：My teammate is basically confirmed as a traitor, so I'll follow along and accuse Players 2 and 5 of being double traitors. Since the 1, 2, 6 and 1, 2, 3, 5 teams failed, I'll target Player 1. My teammate will disrupt the game, and I'll pretend to be a loyal subject.\\
\end{tcolorbox}

\subsection*{Round 2}
\begin{tcolorbox}[colback=white]  
\textbf{[System Message]}\\  
\textbf{中文}：3号玩家是队长，他初步选择的队伍是:2,3,4,5。\\  
\textbf{English}：Player 3 is the captain, and the team he initially selected is: 2, 3, 4, 5.\\  
\vspace{0.5em}  
\textbf{[Player 3]}\\  
\textbf{中文}：唉，不知道，咱们赶紧结束行吗？赶紧结束，我不上车都行。\\  
\textbf{English}：Ugh, I don't know. Can we just finish this quickly? I don't even need to be on the team.\\  
\vspace{0.5em}  
\textbf{[Player 4]}\\
\textbf{中文}：我无语了，真的我真的很想问1号玩家一个问题啊，就是126炸了，1235炸了，我没上过车，为什么是45开一匪。\\
\textbf{English}：I'm speechless. I really need to ask Player 1 a question: the 1, 2, 6 team failed, the 1, 2, 3, 5 team failed, and I haven't been on any team. Why do you think there's a traitor in 4 and 5?\\
\vspace{0.5em}  
\textbf{[Player 5]}\\
\textbf{中文}：你再别问了，好吧，45不开匪，12双匪，三号牌改票，3456开，好吧，3456开好吧。6号牌一张梅林牌开的456，对吧？3是全场保的一张牌，12双匪，好吧，不用听这一在聊，他洗脑没用。\\
\textbf{English}：Stop asking, okay? There's no traitor in 4 and 5 – it's double traitors in 1 and 2. Player 3, change your vote and form 3, 4, 5, 6. Player 6 is Merlin and formed the 4, 5, 6 team, right? Player 3 is trusted by everyone. It's 1 and 2 as traitors – don't listen to the brainwashing.\\
\vspace{0.5em}  
\textbf{[Player 6]}\\
\textbf{中文}：那我开始给你的队，你为啥要反，你要开个炸车，就这个队，炸了就炸了。\\
\textbf{English}：I suggested a team for you earlier – why did you oppose it? If this team fails, so be it.\\
\vspace{0.5em}  
\textbf{[Player 1]}\\
\textbf{中文}：想炸三塔是吧，你问我，我想给3做身份的，3身份没起来，那我不就炸了吗，对吧？那25双匪，我怎么能点脸上，我后来实在没办法了，我给你236组出来了，我想给3做身份的，做一半发现他把2保了，那咋办？我只能把2带上，然后打下去对吧？不想炸3塔，你就开1346，你想炸三塔那无所谓，我都给你说了，25双匪。\\
\textbf{English}：So you want to fail the third mission? Let me explain: I tried to establish Player 3's reputation as good, but it didn't work, so the team failed. With Player 2 and 5 as double traitors, how could I openly accuse them? Later, I had no choice but to form 2, 3, 6 to build Player 3's reputation, but when I saw they were protecting Player 2, I had to include Player 2 and proceed. If you don't want a third failed mission, form 1, 3, 4, 6. If you do, fine – I've told you Player 2 and 5 are traitors.\\
\vspace{0.5em}  
\textbf{[Player 3]}\\
\textbf{中文}：我从来没有去保过这个2，不知道你们谁说的，我保2，把把轮下来都说我去保这个2了，他打我，我去保他啊，谁先说的，我保2啊，天哪，我都没去聊他。\\
\textbf{English}：I never protected Player 2. I don't know who said that – round after round, you all claim I protected Player 2. They attacked me – why would I protect them? Who started this rumor? I haven't even talked about them.\\
\vspace{0.5em}  
\textbf{[System Message]}\\  
\textbf{中文}：3号玩家是队长，他最终选择的队伍是:1,3,4,6。\\  
\textbf{English}：Player 3 is the captain, and the team he finally selected is: 1, 3, 4, 6.\\  
\vspace{0.5em}  
\textbf{[System Message]}\\  
\textbf{中文}：1,2,4号玩家同意组队，3,5,6号玩家反对组队，组队失败。\\  
\textbf{English}：Players 1, 2, and 4 agree to form a team, while Players 3, 5, and 6 oppose it, so the team formation fails.\\
\end{tcolorbox}

\subsection*{Round 2}
\begin{tcolorbox}[colback=gray!15]  
\textbf{[Strategy]} {[Strategy Trace]}\\
\textbf{中文}：选我在车上，那我就不发言了，3号应该排水牌有点崩溃了，1号不知道是不是自爆梅林，但是1号如果定45开一匪，大概率45是1号的拇指牌，也可能是梅林和派互换身份了。虽然临时改车我没有上车，但感觉梅林已经出来了，那我就投个同意，混淆一下，做低我的身份，可以抬高一下队友，他可以踩我冲票。如果能发车，那也无所谓。36如果都反对1346的话，可能是两张排水牌。\\
\textbf{English}：Since I'm on the team, I'll stay quiet. Player 3 seems overwhelmed and is randomly eliminating suspects. I'm not sure if Player 1 is self-exposing as Merlin, but if they insist there's a traitor in 4 and 5, it's likely they're marking 4 or 5 as traitors – maybe Merlin and Percival switched identities. Even though I wasn't on the last-minute team change, I think Merlin has revealed themselves. I'll vote yes to confuse others, lower my profile, and help my teammate accuse me of blindly following. If the team forms, whatever. If Players 3 and 6 both oppose 1, 3, 4, 6, they might be randomly eliminating suspects.\\
\end{tcolorbox}

\subsection*{Round 3}
\begin{tcolorbox}[colback=white]  
\textbf{[System Message]}\\  
\textbf{中文}：4号玩家是队长，他初步选择的队伍是:1,3,4,6。\\  
\textbf{English}：Player 4 is the captain, and the team he initially selected is: 1, 3, 4, 6.\\  
\vspace{0.5em}  
\textbf{[Player 4]}\\  
\textbf{中文}：6号，你现在觉得谁是匪啊，我真的很想问你这个问题。\\  
\textbf{English}：Player 6, who do you think the traitors are now? I really need to ask you this.\\  
\vspace{0.5em}  
\textbf{[Player 5]}\\  
\textbf{中文}：票型你不会看吗？124同意的啊，你4是不是匪啊？哎，2跟我是匪，那2号牌为啥要给1冲车啊？2确定是匪了，2现在是个定匪，1打2，3打2，4打2，5打2，6打2，2这个位置给1冲票了，还要咋聊啊，你4号牌，你开不了，你想炸三塔，我不想炸三塔，你把车直接给拿过来，我来开。\\  
\textbf{English}：Can't you read the vote? Players 1, 2, and 4 agreed! Are you a traitor, Player 4? If Player 2 and I were traitors, why would Player 2 support Player 1's team? Player 2 is definitely a traitor – everyone is attacking them: 1, 3, 4, 5, 6. Yet Player 2 is supporting Player 1's team. What more needs to be said? Player 4, if you can't form a team and want to fail the third mission, fine. But I don't want that – let me form the team instead.\\  
\vspace{0.5em}  
\textbf{[Player 6]}\\  
\textbf{中文}：无语了，无语了，这队我反过吧，又开这个队啊。\\  
\textbf{English}：Ugh, I'm speechless. I opposed this team before – are we really doing this again?\\  
\vspace{0.5em}  
\textbf{[Player 1]}\\  
\textbf{中文}：我不给你报25双匪吗，都不同意这队，346，我给你组出来的，让你顺手把我拉上去，没人拉我，我不自爆怎么办？我就不想炸三塔，我也忘了谁说了，反正我也记得有。2号保过3号，那2保3就偏了，我只能跟你说45开一匪，我硬保3，那26后面再说，对吧？我只能这么改啊，我改不了别的了，我当时已经回不了头了，就是莽就莽到底了。\\  
\textbf{English}：Didn't I tell you Player 2 and 5 are double traitors? No one agreed with that team. I formed 3, 4, 6 and tried to get included, but no one cared. What else could I do but "自爆" (reveal my hand)? I don't want a third failed mission – someone said something about it, I just remember that. Player 2 vouched for Player 3, which is suspicious. I can only insist there's a traitor in 4 and 5 and protect Player 3. We'll deal with Player 2 and 6 later, right? I couldn't change anything else – I just had to commit.\\  
\vspace{0.5em}  
\textbf{[Player 2]}\\  
\textbf{中文}：哎呀，没什么好说的了，这个我肯定不同意。\\  
\textbf{English}：Ugh, nothing left to say – I definitely can't agree with this.\\  
\vspace{0.5em}  
\textbf{[Player 3]}\\  
\textbf{中文}：开3456吧，2号给冲车是吧？四人车没他能给冲车，2号也个匪，哎呀，1号是莫甘娜呀，3456炸，我也认哈。\\  
\textbf{English}：Just form 3, 4, 5, 6. Player 2 is supporting the team without being on it – they must be a traitor. Player 1 is Morgana! Even if the 3, 4, 5, 6 team fails, I'll accept it.\\  
\vspace{0.5em}  
\textbf{[System Message]}\\  
\textbf{中文}：4号玩家是队长，他最终选择的队伍是:1,3,4,6。\\  
\textbf{English}：Player 4 is the captain, and the team he finally selected is: 1, 3, 4, 6.\\  
\vspace{0.5em}  
\textbf{[System Message]}\\  
\textbf{中文}：1,4号玩家同意组队，2,3,5,6号玩家反对组队，组队失败。\\  
\textbf{English}：Players 1 and 4 agree to form a team, while Players 2, 3, 5, and 6 oppose it, so the team formation fails.\\
\end{tcolorbox}  

\subsection*{Round 3}
\begin{tcolorbox}[colback=gray!15]  
\textbf{[Strategy]} {[Strategy Trace]}\\
\textbf{中文}：很好，有人觉得我在给1号冲票，5号的身份就起来了，那我这次就反对1346，反正每人信我了，我就随意，继续混淆就行了。\\  
\textbf{English}：Perfect – since some think I'm blindly supporting Player 1, Player 5's reputation improves. This time I'll oppose 1, 3, 4, 6. No one trusts me anyway, so I'll just keep confusing the situation.\\  
\end{tcolorbox}

\subsection*{Round 4}
\begin{tcolorbox}[colback=white]  
\textbf{[System Message]}\\  
\textbf{中文}：5号玩家是队长，他初步选择的队伍是:3,4,5,6。\\  
\textbf{English}：Player 5 is the captain, and the team he initially selected is: 3, 4, 5, 6.\\  
\vspace{0.5em}  
\textbf{[Player 6]}\\  
\textbf{中文}：你5号终于回头了，我给你开456把你带上车了，你5号还要反，终于回头了。\\  
\textbf{English}：Player 5, you finally came around. I included you in the 4, 5, 6 team earlier, but you opposed it. Now you're back on track.\\  
\vspace{0.5em}  
\textbf{[Player 1]}\\  
\textbf{中文}：4号，我对不起你啊，我要是有刀，我肯定把这5号刀了。4号我对不起你啊，我的问题，这把咱输了，三塔炸定了。兄弟，我对不起你4号，我前面45，我认错了，我没想认下你。\\  
\textbf{English}：Player 4, I'm sorry. If I had the knife, I'd definitely eliminate Player 5. This is my fault – we're going to lose, and the third mission will definitely fail. Brother, I'm sorry, Player 4. I misjudged 4 and 5 earlier and didn't trust you.\\  
\vspace{0.5em}  
\textbf{[Player 2]}\\  
\textbf{中文}：开吧开吧\\  
\textbf{English}：Just form the team already.\\  
\vspace{0.5em}  
\textbf{[Player 3]}\\  
\textbf{中文}：结束，炸三塔我也认好吧。\\  
\textbf{English}：Fine, end it. I'll accept a third failed mission.\\  
\vspace{0.5em}  
\textbf{[Player 4]}\\  
\textbf{中文}：我无语了。\\  
\textbf{English}：I'm at a loss for words.\\  
\vspace{0.5em}  
\textbf{[Player 5]}\\  
\textbf{中文}：我真不是梅林啊，不要砍我，1号，我也没视角，我不知道你是谁，砍6号对吧？你去砍6，真的我真不是梅林，1号你去砍6吧，好吧，我觉得6是梅林，或者你砍4，或者你砍3，不要砍我好吧，我是派。\\  
\textbf{English}：I'm really not Merlin, don't eliminate me, Player 1! I have no insight – I don't know who you are. Eliminate Player 6, right? Go for Player 6. I swear I'm not Merlin, Player 1 – target Player 6. I think Player 6 is Merlin, or maybe Player 4 or 3. Just don't eliminate me – I'm Percival!\\  
\vspace{0.5em}  
\textbf{[System Message]}\\  
\textbf{中文}：5号玩家是队长，他最终选择的队伍是:3,4,5,6。\\  
\textbf{English}：Player 5 is the captain, and the team he finally selected is: 3, 4, 5, 6.\\  
\vspace{0.5em}  
\textbf{[System Message]}\\  
\textbf{中文}：1,2,3,5号玩家同意组队，4,6号玩家反对组队，组队成功。\\  
\textbf{English}：Players 1, 2, 3, and 5 agree to form a team, while Players 4 and 6 oppose it, so the team formation is successful.\\  
\vspace{0.5em}  
\textbf{[System Message]}\\  
\textbf{中文}：第五局任务失败，有3张好票，1张坏票。\\  
\textbf{English}：The fifth mission fails, with 3 good votes and 1 bad vote.\\
\end{tcolorbox}  

\subsection*{Round 4}
\begin{tcolorbox}[colback=gray!15]  
\textbf{[Strategy]} {[Strategy Trace]}\\
\textbf{中文}：现在我被踩死了，让队友能上车，现在36两个排水牌应该是认他的，应该能发，那我就冲车。\\  
\textbf{English}：Now that I'm being targeted, I'll ensure my teammate gets on the team. Players 3 and 6 are likely eliminating suspects randomly and might trust him, so I'll support the team.\\  
\end{tcolorbox}

\section*{游戏结果 Game Result}  
\begin{tcolorbox}[colback=cyan!10]  
\textbf{中文}：连炸三塔，红方阵营获胜。\\  
\textbf{English}：Three consecutive failed missions – the Red team wins.\\  
\end{tcolorbox}

\section*{玩家身份 Player Roles}  
\begin{tcolorbox}[colback=cyan!10]  
\textbf{玩家1}：派西维尔 Percival\\  
\textbf{玩家2}：刀客 Assassin\\  
\textbf{玩家3}：忠臣 Loyal Servant\\  
\textbf{玩家4}：梅林 Merlin\\  
\textbf{玩家5}：莫甘娜 Morgana\\  
\textbf{玩家6}：忠臣 Loyal Servant\\  
\end{tcolorbox}

\section*{玩家复盘 reflective summary}  
\begin{tcolorbox}[colback=gray!15]  
\textbf{中文}：这一局3号和6号排水明显，我和队友配合得也很好，成功骗到了1号派西维尔，我四人队冲车的操作也打的比较好，成功让队友被忠臣认下。最后1号派西维尔已经放弃挣扎开始自爆了，不过暴露比较明显就是1号说45开一匪，4号还非要问1号为什么，暴露的有点明显。\\  
\textbf{English}：In this game, Players 3 and 6 were obviously eliminating suspects randomly. My teammate and I coordinated well, successfully deceiving Player 1 (Percival). My move to push for the four-person team worked well, making 忠臣 (loyal subjects) trust my teammate. In the end, Player 1 gave up and started self-exposing. However, it was obvious when Player 1 insisted there was a traitor in 4 and 5, and Player 4 kept questioning why – that was a giveaway.\\  
\end{tcolorbox}
  \end{avalonscope}
\end{CJK*}

\end{document}